\documentclass[runningheads]{llncs}
\usepackage[T1]{fontenc}
\usepackage{graphicx}
\usepackage{booktabs}
\usepackage{adjustbox}
\usepackage{multirow}
\usepackage{algorithm}
\usepackage{algorithmic}
\usepackage{amsmath}
\usepackage{amssymb}
\usepackage{appendix}
\usepackage{hyperref}
\usepackage[section]{placeins}
\usepackage[draft,commandnameprefix=always]{changes} % use [draft] to show changes
\definechangesauthor[name={Collaborator}, color=red]{COL}
\usepackage{threeparttable}
\usepackage{url}
\usepackage[misc]{ifsym}

\usepackage{hyperref}

\usepackage{mwe}
\begin{document}

\title{CARNet: Cycle-Conditioned Core Aggregation and Redistribution for Multivariate Time Series Forecasting}
\titlerunning{CARNet: Cycle-Conditioned Core Aggregation and Redistribution}
\toctitle{CARNet: Cycle-Conditioned Core Aggregation and Redistribution for Multivariate Time Series Forecasting}
\tocauthor{Awsaf Tausif Adib, Md. Shahria Sarker Shuvo, Md. Estehaar Ahmed Emon, Mustafa Kamal, Fuad Rahman, Shafin Rahman, Nabeel Mohammed}
% If the full title of your paper is short enough to also fit in the running head, you can omit the abbreviated paper title here. You can check as follows: if you comment out the \titlerunning line, something will appear in the header of all odd-numbered pages of your PDF from page 3 onward. This something is either the full title (in which case all is well), or the error message "Title Suppressed Due to Excessive Length". If this error message appears, you're going to want to provide an abbreviated title within the \titlerunning command, because if you won't do it, Springer will do it for you.

%N.B.: Author information (both in the \author{} and \authorrunning{} command) should only be present in the Camera-Ready Version of your paper. The version that you initially submit for review, ought to be double-blind. So, when initially submitting your paper, use:
\author{Awsaf Tausif Adib\inst{1}(\Letter) \and
Md. Shahria Sarker Shuvo\inst{1} \and
Md. Estehaar Ahmed Emon\inst{1} \and
Mustafa Kamal\inst{1} \and
Fuad Rahman\inst{2} \and
Shafin Rahman\inst{1} \and
Nabeel Mohammed\inst{1}}
\authorrunning{A.T. Adib et al.}
\institute{Apurba-NSU R\&D Lab,
Department of Electrical and Computer Engineering,
North South University
\and
Apurba Technologies, California, USA\\
\email{\{awsaf.tasusif, shahria.shuvo, estehaar.emon, mustafa.kamal, shafin.rahman, nabeel.mohammed\}@northsouth.edu}\\
\email{fuad@apurbatech.com}}

% You may leave out the orcidID information, if you want to.
% Use \corr to indicate the corresponding author. Note the spacing around the \corr command. Only one author can be the corresponding author.

%N.B.: comment out the \authorrunning{} command for the double-blind version of your paper submitted for review. Later, if your paper is accepted, use the command for the Camera-Ready Version.
% \authorrunning{A.L. Benjamin et al.}
% If there are two authors, write 'A.L. Benjamin and C.C. Broadus Jr.'
% If there are more than two authors, '[...] et al.' is used.

% \institute{Fictional Southern University, Savannah GA 31404, USA \email{\{a.l.benjamin,a.a.patton\}@fsu.fake}
% \and
% Fictional West Coast University, Long Beach CA 90840, USA \email{ccb@fwcu.fake}
% \and
% Secondary European Affiliation, Tiergartenstr. 17, 69121 Heidelberg, Germany
% \email{lncs@springer.com}}

\maketitle              % typeset the header of the contribution

\begin{abstract}
 Accurately modeling cross-variate dependencies remains a key challenge in multivariate time series forecasting, particularly in the presence of strong periodic patterns. Many existing approaches rely on attention-based mechanisms that incur quadratic complexity and scale poorly with increasing numbers of variates. Recent attention-free aggregation models address this issue through linear-complexity core-based interactions, but they do not explicitly leverage the global periodic structure present in the data. To overcome this limitation, we propose CARNet, a Cycle-Conditioned Core Aggregation and Redistribution framework that integrates global recurrent cycle information into efficient core-based interaction modeling via Multihead Core Aggregation. Extensive experiments on multiple real-world multivariate forecasting benchmarks demonstrate that CARNet consistently outperforms strong transformer and non-attention baselines across diverse prediction horizons while preserving linear-complexity modeling of cross-variate dependencies.
 
\end{abstract}

\section{Introduction}
Accurately modeling cross-variate dependencies is critical for multivariate time series forecasting (MTSF), which underpins a wide range of real-world applications, including energy planning, medical forecasting, climate analysis, and traffic monitoring \cite{tfb,deep_tss_survey,transformers_ts_survey,kedgn}. Recent Transformer-based models that explicitly capture inter-variable interactions, such as iTransformer \cite{itransformer}, have demonstrated clear advantages over channel-independent approaches like PatchTST \cite{patchtst}. However, the quadratic complexity of attention has motivated the development of efficient attention variants \cite{informer,bigbird,linearattention}, as well as attention-free alternatives based on state-space models, graph-based methods, MLP-style architectures and related approaches ~\cite{mamba,smamba,timepro,kedgn,tsmixer,gwn}.

While effective for scalability, most existing channel-dependent models operate directly on raw input series that often exhibit strong periodic or seasonal patterns. Prior work has shown that dominant recurrent structures can obscure other informative dynamics, motivating explicit decomposition strategies or periodicity-aware modeling approaches \cite{autoformer,fedformer}. CycleNet addresses this issue through Residual Cycle Forecasting (RCF), which learns recurrent cycles and removes them from the input to better model residual temporal dynamics \cite{cyclenet}. However, CycleNet primarily focuses on temporal periodicity and does not explicitly model cross-variate dependencies in multivariate settings. TQNet \cite{tqnet} addresses this limitation by introducing recurrent cycles as global correlation representations and conditioning inter-variable interactions through cross-attention, enabling alignment between periodic structures and sample-specific dependencies. While effective, this design relies on attention mechanisms to mediate the interaction between recurrent cycles and variate embeddings. In contrast, efficient attention-free architectures such as SOFTS \cite{softs} and TSMixer \cite{tsmixer} provide scalable alternatives for channel interaction, but they do not explicitly exploit global recurrent cycle information. As a result, integrating cycle-conditioned interaction modeling into attention-free architectures remains non-trivial.

To bridge this gap, we propose CARNet\textsuperscript{\hyperlink{fn:implementation}{1}}, a cycle-conditioned architecture for multivariate time series forecasting that integrates learnable recurrent cycles as global conditioning signals within a unified core aggregation and redistribution framework. \textsc{CARNet} adopts a star-shaped aggregation mechanism~\cite{softs} and introduces an effective way to incorporate global periodic information into cross-variate dependency modeling. Furthermore, to improve the expressiveness of the aggregated representation, we introduce a Multihead Core Aggregation (MHCA) mechanism, which partitions variate representations into multiple interaction subspaces and extracts compact subspace-specific representations through separate head-wise transformations. These subspace representations are then combined to form a unified variate-level representation, which is subsequently pooled across variates to obtain a global cycle-conditioned core. Through Cycle-Conditioned Core Redistribution, this global representation is fed back into the variate embeddings, enabling structured cross-variate interaction modeling while preserving linear complexity and attention-free computation.

In summary, the contributions of this paper are as follows:

\begin{itemize}
    \item[1]. We propose \textbf{CARNet}, a cycle-conditioned architecture for multivariate time series forecasting that incorporates learnable recurrent cycles as global conditioning signals within a unified aggregation and redistribution framework.

    \item[2]. We introduce a \emph{Multihead Core Aggregation (MHCA)} mechanism that enhances cross-variate interaction by extracting and combining multiple core representations from partitioned variate features.

    \item[3]. We conduct extensive experiments on 12 real-world multivariate time series benchmarks, demonstrating that CARNet consistently achieves strong performance across multiple forecasting horizons with favorable efficiency.
\end{itemize}
\footnotetext[1]{\hypertarget{fn:implementation}{}Official implementation: \url{https://github.com/adib3552/carnet}}

\section{Related Work}
\subsection{Aggregation Models}
Aggregation-based forecasting models aim to capture cross-variate dependencies by compressing information from multiple variables into shared latent representations and then redistributing the aggregated information for prediction. Compared with attention-based methods, these models often provide a more efficient alternative for multivariate forecasting by avoiding explicit pairwise interaction across all variables. Earlier lightweight architectures such as TSMixer~\cite{tsmixer} and TiDE~\cite{tide} explored efficient channel interaction through mixing and MLP-based designs, demonstrating that strong forecasting performance can be achieved without relying on full attention mechanisms.

\begin{figure*}[t]
\centering
\includegraphics[width=\textwidth]{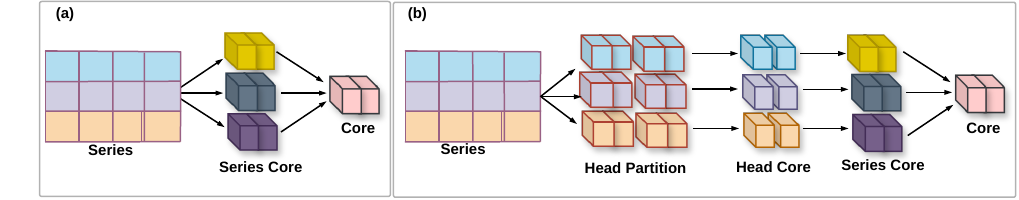}
\caption{Comparison of core aggregation strategies. (a) The traditional approach aggregates all variates into a single shared core representation. (b) Our Multihead Core Aggregation first partitions variates into multiple interaction subspaces, extracts head-specific cores, and then combines them into a shared core, enabling richer and more diverse cross-variate interaction modeling.}
\label{Aggregation}
\end{figure*}

More recently, SOFTS~\cite{softs} introduced a dedicated core aggregation mechanism that summarizes multivariate information into a compact shared core and redistributes it back to each variate, achieving strong performance with linear complexity. As illustrated in Figure~\ref{Aggregation}, our work builds on this aggregation paradigm by moving beyond single-core aggregation and introducing Multihead Core Aggregation, which captures richer and more diverse cross-variate interactions. This design enables CARNet to preserve the efficiency advantages of aggregation-based modeling while enhancing its representational flexibility.
\subsection{Periodic Models}

Periodic models explicitly capture recurring temporal patterns by introducing inductive biases toward seasonality, cycles, and frequency-domain structures. 
Autoformer~\cite{autoformer} popularized this paradigm by decomposing input sequences into periodic (seasonal) and non-periodic (trend) components, allowing models to focus on structured temporal regularities. 
Similarly, FEDformer~\cite{fedformer} leverages frequency-domain representations together with attention mechanisms to model global periodic dependencies for long-horizon forecasting.

Several lightweight architectures further exploit periodic structures through efficient temporal modeling. 
TimeMixer~\cite{timemixer} captures periodic patterns using multi-scale mixing across different temporal resolutions, while SparseTSF~\cite{sparsetsf} introduces sparse temporal interactions to selectively model dominant periodic dependencies. 
TQNet~\cite{tqnet} models periodicity using quantized temporal representations, enabling efficient learning of recurring patterns across multiple temporal scales. 
Together, these methods demonstrate that explicitly modeling periodic structures can simplify temporal dynamics and improve forecasting performance.

\subsection{Channel-Dependent Modeling}

Modeling cross-variate dependencies is central to multivariate time series forecasting, since correlations across variables often improve prediction accuracy. Transformer-based methods address this through explicit channel-aware representations. For instance, iTransformer~\cite{itransformer} treats variables as tokens to model cross-variate relations via self-attention, while TimeXer~\cite{timexer} introduces global learnable tokens to summarize channel-specific temporal information and support inter-channel interaction. TimesNet~\cite{timesnet} also captures cross-variate dependencies through shared representations while emphasizing structured temporal pattern extraction.

Beyond attention-based architectures, several lightweight approaches provide efficient alternatives for channel interaction modeling. TSMixer~\cite{tsmixer} uses alternating temporal and channel mixing layers in an MLP-based framework, while SOFTS~\cite{softs} introduces a core aggregation mechanism that summarizes information across variables into a shared latent representation and redistributes it back to each channel. Together, these methods reflect diverse strategies for modeling cross-variate dependencies in multivariate forecasting.

\section{Methodology}
Multivariate time series forecasting aims to predict a future sequence $\mathbf{Y} = [\mathbf{y}_{T+1}, \mathbf{y}_{T+2}, \ldots, \mathbf{y}_{T+H}] \in \mathbb{R}^{C \times H}$ given a historical input sequence of the past $\mathbf{X} = [\mathbf{x}_1, \mathbf{x}_2, \ldots, \mathbf{x}_T] \in \mathbb{R}^{C \times T}$, where $C$ denotes the number of variates (channels), $T$ is the look-back window length, and $H$ is the forecasting horizon. Formally, the objective is to learn a parametric forecasting model $f{\boldsymbol{\phi}}(\cdot)$ with learnable parameters $\boldsymbol{\phi}$ that maps the past window to the future window, i.e., $\mathbf{Y} = f{\boldsymbol{\phi}}(\mathbf{X})$, where $f{\boldsymbol{\phi}}: \mathbb{R}^{C \times T} \rightarrow \mathbb{R}^{C \times H}$. The forecasting function is designed to capture cross-variate dependencies while accounting for recurring temporal patterns shared across variates. In many real-world scenarios, such dependencies are influenced by underlying periodic structures, motivating the incorporation of global cycle information into variate interaction modeling.

\subsection{Overview of CARNet}

\begin{figure*}[t]
\centering
\includegraphics[width=\textwidth]{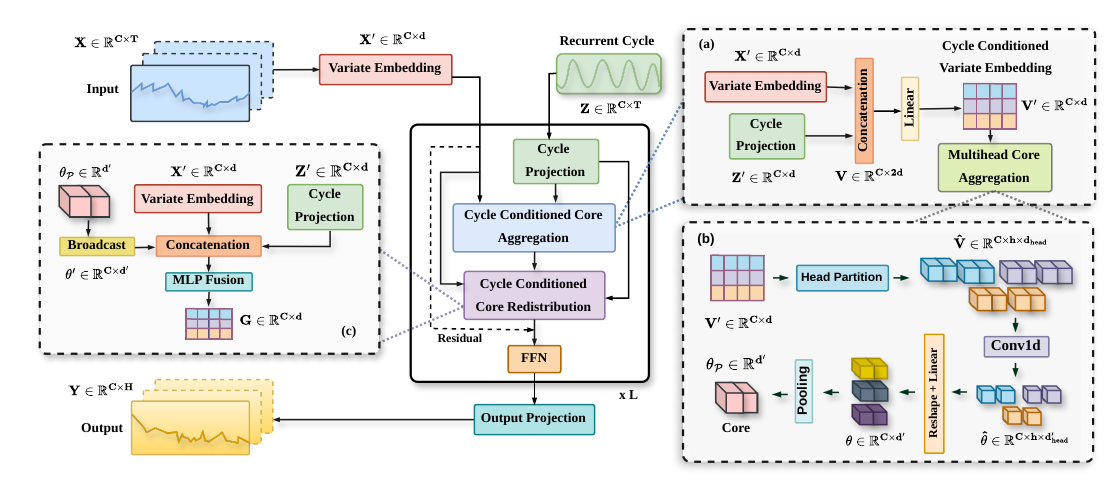}
\caption{Overview of the CARNet framework. 
(a) In the Cycle-Conditioned Core Aggregation module, the variate embedding and cycle projection are fused to obtain a cycle-conditioned variate embedding. 
(b) The Multihead Core Aggregation module extracts a compact core representation from the cycle-conditioned variate embedding. 
(c) In the Cycle-Conditioned Core Redistribution module, the shared core is redistributed to the variate embeddings together with the cycle projection.}
\label{CARNet}
\end{figure*}

\noindent As illustrated in Figure~\ref{CARNet}, CARNet follows a Cycle-Conditioned Core Aggregation and Redistribution pipeline for multivariate time series forecasting. Given an input multivariate time series $\mathbf{X} \in \mathbb{R}^{C \times T}$, where $C$ denotes the number of variates and $T$ the look-back length, CARNet first constructs a learnable recurrent cycle representation $\mathbf{Z} \in \mathbb{R}^{C \times T}$ to encode dominant periodic patterns shared across variates. This cycle representation is phase-aligned with the input sequence and serves as a global conditioning signal. Since CARNet is explicitly channel-dependent, the raw input series is transformed into a variate-level embedding $\mathbf{X}' \in \mathbb{R}^{C \times d}$ through a linear embedding layer. In parallel, the cycle representation is projected into the same embedding space, yielding $\mathbf{Z}' \in \mathbb{R}^{C \times d}$. The two representations are concatenated along the feature dimension to form $\mathbf{V} \in \mathbb{R}^{C \times 2d}$ and linearly projected to obtain a cycle-conditioned variate representation $\mathbf{V}' \in \mathbb{R}^{C \times d}$. To model cross-variate interactions, the cycle-conditioned variate embedding $\mathbf{V}'$ is passed to the Multihead Core Aggregation (MHCA) module to extract a compact variate-wise core representation. As shown in Figure~\ref{CARNet}  MHCA partitions the embedding dimension into $h$ heads and extracts a compact representation from each subspace. The resulting head-wise representations are combined to form a unified Multihead representation $\boldsymbol{\theta} \in \mathbb{R}^{C \times d'}$. Finally, the core $\boldsymbol{\theta}_{\mathcal{P}} \in \mathbb{R}^{d'}$ is obtained through stochastic pooling across the variate dimension. The Cycle-Conditioned Core is then broadcast to each channel to obtain $\boldsymbol{\theta}' \in \mathbb{R}^{C \times d'}$ and redistributed to the variate embedding along with the cycle projection using MLP fusion to produce refined representations $\mathbf{G} \in \mathbb{R}^{C \times d}$, making the redistribution cycle-conditioned. Finally, the refined representation is added to the variate embedding through a residual connection and passed through a Feed-Forward Network (FFN), after which the output is projected to the forecasting horizon to generate the multivariate prediction $\mathbf{Y} \in \mathbb{R}^{C \times H}$.

Overall, CARNet integrates periodic structure into attention-free cross-variate modeling through structured core aggregation and redistribution, enabling efficient and scalable forecasting.

\subsection{Components of CARNet}
\noindent \textbf{Learnable Periodic Patterns.}
Following CycleNet~\cite{cyclenet} and the temporal query (TQ)
mechanism of TQNet~\cite{tqnet}, we adopt a learnable recurrent cycle
representation $\mathbf{Q} \in \mathbb{R}^{C \times W}$, where $C$
denotes the number of channels and $W$ is the maximum stable cycle
length estimated using autocorrelation analysis (ACF)~\cite{madsen_ts}.
The cycle matrix $\mathbf{Q}$ is initialized to zero and jointly
optimized with the forecasting model via backpropagation. For an input sequence starting at time index $t$, the phase-aligned cyclic
sequence $\mathbf{Z} \in \mathbb{R}^{C \times T}$ is defined at each
look-back index $\ell = 0,\dots,T-1$ as 
\begin{equation}
\mathbf{Z}_{:,\,\ell}
=
\mathbf{Q}_{:,\, (t+\ell)\bmod W},
\quad \ell = 0,\dots,T-1 .
\end{equation}

\noindent\textbf{Cycle-Conditioned Core Aggregation.}
Channel-dependent models such as iTransformer~\cite{itransformer}
and SOFTS~\cite{softs} first project each variate into a latent
representation of dimension $d$. Following this paradigm, we obtain a
variate-level embedding $\mathbf{X}' \in \mathbb{R}^{C \times d}$ from
the raw input sequence and project the learnable recurrent cycle into
the same latent space to obtain
$\mathbf{Z}' \in \mathbb{R}^{C \times d}$. Here, $\mathbf{X}'$
encodes local input-level variate information, while
$\mathbf{Z}'$ encodes the global periodic structure shared across the
training set~\cite{tqnet}. The two representations are concatenated
along the feature dimension, followed by linear projection and nonlinear activation, yielding the cycle-conditioned variate embedding $V'$
\begin{equation}
\mathbf{V}' =
\psi\!\left(
\mathcal{L}_V\big(\mathcal{F_{A}}(\mathbf{X}', \mathbf{Z}')\big)
\right)
\in \mathbb{R}^{C \times d}.
\end{equation}
Here, $\mathcal{F_{A}}(\cdot,\cdot)$ denotes feature-wise concatenation,
$\mathcal{L}_V:\mathbb{R}^{C \times 2d}\rightarrow
\mathbb{R}^{C \times d}$ denotes linear projection and
$\psi(\cdot)$ denotes the GELU activation function~\cite{gelu}.

To model cross-variate interaction under periodic conditioning, we first
extract a compact shared core representation through a Multihead Core
Aggregation module, where the cycle-conditioned variate embedding
$\mathbf{V}' \in \mathbb{R}^{C \times d}$ is partitioned into $h$ heads
along the feature dimension using a linear projection followed by a
reshaping function, producing
$\mathbf{\hat{V}} \in \mathbb{R}^{C \times h \times d_{\text{head}}}$,
where $d_{\text{head}} = d/h$.
Each head is then independently transformed into a per-head core
dimension $d'_{\text{head}}$ through a grouped $1$D convolution,
producing the head-wise feature representation
$\hat{\boldsymbol{\theta}} \in \mathbb{R}^{C \times h \times d'_{\text{head}}}$.
The $i$-th head-wise feature is denoted by
$\hat{\boldsymbol{\theta}}_i \in \mathbb{R}^{C \times d'_{\text{head}}}$,
where $d'_{\text{head}} = d'/h$ and $i=1,\dots,h$.
The head-wise features are then reshaped and refined via a linear layer
to obtain the unified Multihead representation
$\boldsymbol{\theta} \in \mathbb{R}^{C \times d'}$.
Pooling is then applied across the variate dimension to produce a shared
compact core $\boldsymbol{\theta}_{\mathcal{P}} \in \mathbb{R}^{d'}$,
which is later utilized to condition the variate embeddings by
broadcasting it across all channels. The overall aggregation process can be compactly expressed as
\begin{equation}
\begin{aligned}
\boldsymbol{\theta} &=
\mathcal{L}_{\theta}\!\Big(
\alpha\!\Big(
\big[
\psi(\omega(\mathbf{\hat{V}}_1)),\;
\dots,\;
\psi(\omega(\mathbf{\hat{V}}_h))
\big]
\Big)
\Big)
\in \mathbb{R}^{C \times d'},\\
\boldsymbol{\theta}' &=
\beta\!\left(\mathcal{P}(\boldsymbol{\theta})\right)
\in \mathbb{R}^{C \times d'}.
\end{aligned}
\end{equation}

Here, $\alpha(\cdot)$ denotes the reshaping function,
$\mathcal{P}(\cdot)$ denotes pooling across the variate dimension,
specifically stochastic pooling~\cite{stochasticpooling} which was also used in prior aggregation work~\cite{softs}, $\omega(\cdot)$ denotes a
$1$D convolution with kernel size $1$ and
$\mathcal{L}_{\theta}(\cdot)$ denote
linear layer applied along the feature dimension.
The operator $\beta(\cdot)$ broadcasts the pooled core across all
variates and $\psi(\cdot)$ denotes the GELU activation function.

\noindent\textbf{Cycle-Conditioned Core Redistribution.}
To enable interaction between global periodic structure and local variate representations, the shared core $\theta'$ together with the cycle projection $\mathbf{Z'}$ are redistributed into the variate embedding. The redistribution process begins by concatenating the variate embedding with the cycle projection and core along the feature dimension. The fused representation is then passed through a MLP block consisting of two linear layers to obtain a refined representation $\mathbf{G} \in \mathbb{R}^{C \times d}$. The refined representation is added to the original variate embedding via a residual connection, passed through a Feed-Forward Network (FFN), producing the output representation $\mathbf{O} \in \mathbb{R}^{C \times d}$. The redistribution process can be compactly expressed as
\begin{equation}
\begin{aligned}
\mathbf{G}
&=
\mathcal{L}_F\!\Big(
\psi\big(
\mathcal{L}_{\tilde{V}}
(
\mathcal{F_{R}}(\mathbf{X'}, \mathbf{Z'},\boldsymbol{\theta}')
)
\big)
\Big),\\
\mathbf{O}
&=
\mathrm{FFN}\!\left(
\mathbf{X}' + \mathbf{G}
\right).
\end{aligned}
\end{equation}

Here, $\mathcal{F_{R}}(\cdot,\cdot,\cdot)$ denotes feature-wise concatenation of the variate embedding, cycle projection, and core along the feature dimension. The linear layer $\mathcal{L}_{\tilde{V}} : \mathbb{R}^{C \times (2d+d')} \rightarrow \mathbb{R}^{C \times d}$ projects the fused representation into the latent feature space, while the final linear layer $\mathcal{L}_F$ further refines the redistributed features. The operator $\mathrm{FFN}(\cdot)$ denotes a convolutional feed-forward network applied independently across variates.

\noindent\textbf{Output Projection.}
The final representation is mapped to the forecasting horizon through
a linear transformation applied along the feature dimension:
\begin{equation}
\mathbf{Y}
=
\mathcal{L}_O(\mathbf{O})
\in \mathbb{R}^{C \times H}.
\end{equation}

The operator $\mathcal{L}_O :
\mathbb{R}^{C \times d}
\rightarrow
\mathbb{R}^{C \times H}$
denotes a linear projection that transforms latent representations
into the forecasting horizon. A step-by-step pseudocode description is provided in Appendix \ref{app:Algorithm} for clarity and reproducibility.

\subsection{Loss Function } 
\noindent We employ the Mean Squared Error (MSE) loss as the optimization objective, which is the standard choice for regression-based time series forecasting. Given the model predictions $\hat{Y}$ and the ground truth targets $Y$, the loss is defined as:
\begin{equation}
    \mathcal{L}_{\text{MSE}} = \frac{1}{N} \sum_{i=1}^{N} (\hat{Y}_i - Y_i)^2.
\end{equation}

\subsection{Normalization}
To ensure stable training across variables with different scales, we apply
Instance Normalization \cite{instancenorm} to the input sequence. For each channel $c$, the
input is normalized across the temporal dimension and later restored to
its original scale through de-normalization as
\begin{equation}
\begin{aligned}
\hat{X}_{b,t,c} &=
\frac{X_{b,t,c} - \mu_{b,c}}
{\sqrt{\sigma_{b,c}^{2} + \epsilon}}, \\
Y_{b,\tau,c} &=
\hat{Y}_{b,\tau,c}\sqrt{\sigma_{b,c}^{2} + \epsilon} + \mu_{b,c},
\quad \tau = 1,\dots,H.
\end{aligned}
\end{equation}
Here, $\mu_{b,c}$ and $\sigma_{b,c}$ denote the mean and standard deviation
computed over the temporal dimension for each channel, and $\epsilon$ is
a small constant for numerical stability.

\subsection{Time Complexity Analysis.}
CARNet avoids pairwise attention and therefore does not incur quadratic complexity with respect to either the number of variates $C$ or the sequence length $T$. Constructing the cyclic sequence costs $\mathcal{O}(CT)$, while variate embedding and cycle projection require $\mathcal{O}(CTd)$ and $\mathcal{O}(Cd^{2})$, respectively. In MHCA, projection into interaction subspaces costs $\mathcal{O}(Cd^{2})$, head-wise core extraction via grouped $1$D convolution costs $\mathcal{O}(Cdd')$, cross-head mixing adds $\mathcal{O}(C(d')^{2})$, and stochastic pooling contributes $\mathcal{O}(Cd')$. During redistribution, fusion and projection require $\mathcal{O}(C(2d+d')d)$, followed by refinement with $\mathcal{O}(Cd^{2})$, while the convolutional feed-forward network adds $\mathcal{O}(Cdd_{\text{ff}})$. Therefore, the per-layer time complexity of CARNet is $\mathcal{O}\!\left(CTd + Cd^{2} + Cdd' + C(d')^{2} + Cdd_{\text{ff}}\right)$ which scales linearly with both $C$ and $T$.

\section{Experiments and Results}

\subsection{Setup}
\noindent \textbf{Implementation Details.} All experiments are implemented in PyTorch \cite{pytorch}. Models are trained using the Adam optimizer \cite{adam} with a One-Cycle learning rate scheduler. Hyperparameters are selected via grid search. We tune the hidden dimension $d$, core dimension $d'$, feed-forward dimension $d_{\text{ff}}$, number of encoder layers $e_{\text{layers}}$, and batch size. Specifically, we consider $d, d_{\text{ff}} \in \{128, 256, 512\}$, batch sizes in $\{16, 32\}$, and encoder layers $e_{\text{layers}} \in \{1, 2, 3, 4\}$. The core dimension $d'$ is constrained to satisfy $d' \leq d$ to maintain a compact core representation. These hyperparameter choices follow settings commonly adopted in prior core-based work \cite{softs}. The cycle length $W$ and learning rate used for each dataset are reported in Table~\ref{tab:dataset_stats}. The cycle length $W$ is determined following CycleNet~\cite{cyclenet} and  TQNet~\cite{tqnet}, where it is estimated via autocorrelation analysis  (ACF) and shown through ablation studies to reflect intrinsic dataset periodicity rather than a tunable model hyperparameter.  We therefore fix $W$ to the reported values for fair comparison and do not perform additional cycle-length search. Additional experimental details are provided in Appendix C.

\begin{table}[!h]
\caption{Detailed information about the datasets. 
The cycle length $W$ follows the values reported in CycleNet~\cite{cyclenet} 
and TQNet~\cite{tqnet}, where it is determined via autocorrelation analysis (ACF). We also report the learning rate used for each dataset.}
\label{tab:dataset_stats}
\centering
\footnotesize
\begin{adjustbox}{max width=\columnwidth}
\setlength{\tabcolsep}{9pt}
\renewcommand{\arraystretch}{0.5}
\begin{tabular}{l c c c c c l}
\toprule
Dataset & Channels & Timesteps & Interval & Cycle Length $W$ & Learning Rate & Domain \\
\midrule
ETTh1       & 7   & 14,400 & 1 hour  & 24  & $1\times10^{-4}$ & Electricity \\
ETTh2       & 7   & 14,400 & 1 hour  & 24  & $1\times10^{-4}$ & Electricity \\
ETTm1       & 7   & 57,600 & 15 mins & 96  & $1\times10^{-4}$ & Electricity \\
ETTm2       & 7   & 57,600 & 15 mins & 96  & $1\times10^{-4}$ & Electricity \\
Electricity & 321 & 26,304 & 1 hour  & 168 & $1\times10^{-3}$ & Electricity \\
Traffic     & 862 & 17,544 & 1 hour  & 168 & $1\times10^{-3}$ & Transportation \\
Weather     & 21  & 52,696 & 10 mins & 144 & $1\times10^{-4}$ & Weather \\
Solar     & 137  &  52,560 & 10 mins & 144 & $1\times10^{-4}$ & Energy \\
PEMS03     & 358  &  26,208 & 5 mins & 288 & $1\times10^{-3}$ & Transportation \\
PEMS04     & 307  & 16,992 & 5 mins & 288 & $1\times10^{-3}$ & Transportation \\
PEMS07     & 883  & 28,224 & 5 mins & 288 & $1\times10^{-3}$ & Transportation \\
PEMS08     & 170  &  17,856 & 5 mins & 288 & $1\times10^{-3}$ & Transportation \\

\bottomrule
\end{tabular}
\end{adjustbox}
\end{table}

\begin{table}[!h]
  \caption{Multivariate forecasting results on 12 real-world benchmark datasets under prediction horizons $H \in \{96, 192, 336, 720\}$ and $H \in \{12, 24, 48, 96\}$ with a fixed look-back window $T = 96$. Baseline results are adopted from the standardized benchmark evaluations reported in TQNet~\cite{tqnet} and SOFTS~\cite{softs}, which include comparisons with all competing methods under identical experimental settings. The best results are shown in \textbf{\textcolor{red}{bold}} and the second-best results are shown in \underline{\textcolor{blue}{underline}}.}

  \centering
  %\footnotesize
%\normalfont
%\setlength{\tabcolsep}{1pt}
\renewcommand{\arraystretch}{0.95}
  \resizebox{1\columnwidth}{!}{
{
  % \begin{threeparttable}
  % \begin{small}
  % \renewcommand{\multirowsetup}{\centering}
  % \setlength{\tabcolsep}{1.45pt}
  \label{tab:detailed_res}
  % \scalebox{0.4}{
  \begin{tabular}{c|c|cc|cc|cc|cc|cc|cc|cc|cc|cc|cc|cc|cc}
    \toprule
    \multicolumn{2}{c}{Models} & 
    \multicolumn{2}{c}{CARNet (ours)}   &
    \multicolumn{2}{c}{TQnet \cite{tqnet}} & 
    \multicolumn{2}{c}{CycleNet \cite{cyclenet}} & 
    \multicolumn{2}{c}{TimeXer \cite{timemixer}}   &
    \multicolumn{2}{c}{SOFTS \cite{softs}} & 
    \multicolumn{2}{c}{iTransformer \cite{itransformer}} & 
    \multicolumn{2}{c}{Crossformer \cite{crossformer}} & 
    \multicolumn{2}{c}{TiDE \cite{tide}} & 
    \multicolumn{2}{c}{SCINet \cite{scinet}} & 
    \multicolumn{2}{c}{DLinear \cite{dlinear}} \\
    \cmidrule(lr){3-4} \cmidrule(lr){5-6}\cmidrule(lr){7-8} \cmidrule(lr){9-10}\cmidrule(lr){11-12}\cmidrule(lr){13-14} \cmidrule(lr){15-16} \cmidrule(lr){17-18} \cmidrule(lr){19-20} \cmidrule(lr){21-22}\cmidrule(lr){23-24}
    \multicolumn{2}{c}{Metric} & MSE & MAE & MSE & MAE & MSE & MAE & MSE  & MAE & MSE & MAE & MSE & MAE & MSE & MAE & MSE & MAE & MSE & MAE & MSE & MAE \\
    \midrule\multirow{5}{*}{\rotatebox{90}{ETTm1}}
    & 96	& \textbf{\textcolor{red}{0.309}}&	\textbf{\textcolor{red}{0.352}}&	\underline{\textcolor{blue}{0.311}}&	\underline{\textcolor{blue}{0.353}}&	0.319&	0.360	&0.318&	0.356&	0.325	&0.368	&0.334&	0.368	&0.404	&0.426	&0.364	&0.387&	0.418	&0.438	&0.345	&0.372\\ 
    &192&	\textbf{\textcolor{red}{0.354}}&	\underline{\textcolor{blue}{0.379}}&	\underline{\textcolor{blue}{0.356}}	&\textbf{\textcolor{red}{0.378}}&	0.360	&0.381&	0.362&	0.383&	0.375&	0.391&	0.377&	0.391	&0.450	&0.451	&0.398&	0.400	&0.439	&0.450&	0.380	&0.389	\\ 
   & 336	&\textbf{\textcolor{red}{0.388}}&	\textbf{\textcolor{red}{0.401}}	&0.390&	\textbf{\textcolor{red}{0.401}}	&\underline{\textcolor{blue}{0.389}}	&\underline{\textcolor{blue}{0.403}}&	0.395&	0.407&	0.405&	0.420&	0.426&	0.420&	0.532&	0.515	&0.428	&0.425&	0.490&	0.485&	0.413&	0.413\\ 
    & 720	&0.454&	\textbf{\textcolor{red}{0.438}}&	\underline{\textcolor{blue}{0.452}}&	\underline{\textcolor{blue}{0.440}}&	\textbf{\textcolor{red}{0.447}}&	0.441&	0.452&	0.441	&0.466	&0.459	&0.491&	0.459&	0.666&	0.589	&0.487	&0.461	&0.595	&0.550	&0.474	&0.453\\ 
    \cmidrule(lr){2-24}  & Avg&	\textbf{\textcolor{red}{0.376}}&	\textbf{\textcolor{red}{0.393}}	&\underline{\textcolor{blue}{0.377}}	&\underline{\textcolor{blue}{0.393}}&	0.379&	0.396	&0.382	&0.397	&0.393&	0.410	&0.407&	0.410	&0.513&	0.495&	0.419&	0.418&	0.486	&0.481	&0.419	&0.424\\ 
    \midrule\multirow{5}{*}{\rotatebox{90}{ETTm2}} 
    & 96	&\underline{\textcolor{blue}{0.171}}&	\underline{\textcolor{blue}{0.254}}&	0.173&	0.256&	\textbf{\textcolor{red}{0.163}}&	\textbf{\textcolor{red}{0.246}}	&\underline{\textcolor{blue}{0.171}}	&0.256	&0.180&	0.264&	0.180&	0.264	&0.287	&0.366	&0.207	&0.305	&0.286	&0.377	&0.193	&0.292\\ 
    & 192&	0.239&	\underline{\textcolor{blue}{0.298}}&	0.238	&\underline{\textcolor{blue}{0.298}}	&\textbf{\textcolor{red}{0.229}}&	\textbf{\textcolor{red}{0.290}}&	\underline{\textcolor{blue}{0.237}}	&0.299&	0.246&	0.309&	0.250&	0.309&	0.414&	0.492&	0.290&	0.364	&0.399	&0.445&	0.284	&0.362	\\ 
    & 336	&0.298&	\underline{\textcolor{blue}{0.338}}	&0.301	&0.340	&\textbf{\textcolor{red}{0.284}}&	\textbf{\textcolor{red}{0.327}}&	\underline{\textcolor{blue}{0.296}}	&\underline{\textcolor{blue}{0.338}}	&0.319	&0.348	&0.311	&0.348	&0.597	&0.542	&0.377	&0.422&	0.637&	0.591	&0.369&	0.427\\ 
    & 720	&0.404	&0.400	&0.397	&0.396&	\textbf{\textcolor{red}{0.389}}	&\textbf{\textcolor{red}{0.391}}	&\underline{\textcolor{blue}{0.392}}	&\underline{\textcolor{blue}{0.394}}	&0.405	&0.407	&0.412	&0.407	&1.730	&1.042	&0.558	&0.524	&0.960	&0.735	&0.554	&0.522\\ 
    \cmidrule(lr){2-24}  & Avg	&0.278	&0.323&	0.277	&0.323	&\textbf{\textcolor{red}{0.266}}	&\textbf{\textcolor{red}{0.314}}	&\underline{\textcolor{blue}{0.274}}	&\underline{\textcolor{blue}{0.322}}	&0.287	&0.332&	0.288	&0.332	&0.757	&0.610	&0.358	&0.404	&0.571	&0.537	&0.350	&0.401\\ 
    \midrule\multirow{5}{*}{\rotatebox{90}{ETTh1}} 
    & 96	&\underline{\textcolor{blue}{0.373}}	&0.396	&\textbf{\textcolor{red}{0.371}}	&\textbf{\textcolor{red}{0.393}}	&0.375	&\underline{\textcolor{blue}{0.395}}	&0.382	&0.403	&0.381	&0.405	&0.386&	0.405	&0.423	&0.448	&0.479	&0.464	&0.654	&0.599	&0.386	&0.400\\ 
    & 192&	\textbf{\textcolor{red}{0.425}}	&\textbf{\textcolor{red}{0.424}}	&0.428	&\underline{\textcolor{blue}{0.426}}&	\underline{\textcolor{blue}{0.427}}&	0.428&	0.429	&0.435	&0.435	&0.436&	0.441&	0.436&	0.471&	0.474	&0.525	&0.492	&0.719	&0.631	&0.437	&0.432\\ 
    & 336	&\textbf{\textcolor{red}{0.466}}	&\underline{\textcolor{blue}{0.446}}	&0.476	&\underline{\textcolor{blue}{0.446}}	&0.472	&\textbf{\textcolor{red}{0.409}}&	\underline{\textcolor{blue}{0.468}}	&0.448&	0.480	&0.458&	0.487&	0.458&	0.570&	0.546&	0.565	&0.515	&0.778	&0.659	&0.481	&0.459\\ 
    & 720	&\underline{\textcolor{blue}{0.471}}	&0.471	&0.487	&\underline{\textcolor{blue}{0.470}}	&0.476&	0.520&	\textbf{\textcolor{red}{0.469}}&	\textbf{\textcolor{red}{0.461}}&	0.499&	0.491&	0.503&	0.491&	0.653&	0.621&	0.594	&0.558&	0.836	&0.699	&0.519	&0.516\\ 
    \cmidrule(lr){2-24}  & Avg&	\textbf{\textcolor{red}{0.434}}&	\textbf{\textcolor{red}{0.434}}	&0.441	&\textbf{\textcolor{red}{0.434}}&	0.438&	0.438&	\underline{\textcolor{blue}{0.437}}	&\underline{\textcolor{blue}{0.437}}&	0.449&	0.447&	0.454	&0.447&	0.529&	0.522&	0.541&	0.507&	0.747	&0.647&	0.456	&0.452\\ 
    \midrule\multirow{5}{*}{\rotatebox{90}{ETTh2}} 
    & 96&	\underline{\textcolor{blue}{0.295}}	&0.346	&\underline{\textcolor{blue}{0.295}}&	\underline{\textcolor{blue}{0.343}}	&0.298&	0.344	&\textbf{\textcolor{red}{0.286}}&	\textbf{\textcolor{red}{0.338}}&	0.297&	0.349&	0.297&	0.349&	0.745&	0.584&	0.400	&0.440	&0.707	&0.621&	0.333&	0.387\\ 
    & 192&	0.376&	0.397&	\underline{\textcolor{blue}{0.367}}&	\underline{\textcolor{blue}{0.393}}&	0.372&	0.396&	\textbf{\textcolor{red}{0.363}}&	\textbf{\textcolor{red}{0.389}}	&0.373&	0.400&	0.380&	0.400&	0.877&	0.656&	0.528	&0.509	&0.860	&0.689&	0.477	&0.476\\ 
    & 336	&0.423&	0.434&	0.417	&\underline{\textcolor{blue}{0.427}}&	0.431&	0.439&\underline{\textcolor{blue}{0.414}}	&\textbf{\textcolor{red}{0.423}}	&\textbf{\textcolor{red}{0.410}}&	0.432	&0.428&	0.432&	1.043	&0.731	&0.643&	0.571	&1.000&	0.744	&0.594&	0.541\\ 
    & 720	&0.449	&0.458	&0.433	&0.446&	0.450&	0.458&	\textbf{\textcolor{red}{0.408}}&	\textbf{\textcolor{red}{0.432}}&	\underline{\textcolor{blue}{0.411}}&	\underline{\textcolor{blue}{0.445}}&	0.427&	0.445&	1.104&	0.763&	0.874&	0.679&	1.249&	0.838&	0.831&	0.657\\
    \cmidrule(lr){2-24}  & Avg	&0.386	&0.409	&0.378	&\underline{\textcolor{blue}{0.402}}&	0.388	&0.409	&\textbf{\textcolor{red}{0.368}}&	\textbf{\textcolor{red}{0.397}}	&\underline{\textcolor{blue}{0.373}}&	0.407	&0.383	&0.407&	0.942	&0.684	&0.611&	0.550	&0.954&	0.723&	0.559&	0.515\\
    \midrule\multirow{5}{*}{\rotatebox{90}{ECL}}  
    & 96	&\textbf{\textcolor{red}{0.131}}&	\textbf{\textcolor{red}{0.226}}	&\underline{\textcolor{blue}{0.134}}	&\underline{\textcolor{blue}{0.229}}&	0.136&	\underline{\textcolor{blue}{0.229}}&	0.140&	0.242&	0.143&	0.233&	0.148&	0.240 &	0.219&	0.314&	0.237&	0.329&	0.247&	0.345&	0.197&	0.282\\ 
    & 192	&\textbf{\textcolor{red}{0.149}}	&\textbf{\textcolor{red}{0.244}}&	0.154	&\underline{\textcolor{blue}{0.247}}&	\underline{\textcolor{blue}{0.152}}&	\textbf{\textcolor{red}{0.244}}&	0.157&	0.256&	0.158 &	0.248 &0.162 &	0.253&	0.231&	0.322&	0.236	&0.330&	0.257	&0.355	&0.196&	0.285\\ 
    &336	&\textbf{\textcolor{red}{0.164}}	&\textbf{\textcolor{red}{0.260}}	&\underline{\textcolor{blue}{0.169}}	&\underline{\textcolor{blue}{0.264}}	&0.170	&\underline{\textcolor{blue}{0.264}}&	0.176&	0.275	&0.178&	0.269&	0.178&	0.269&	0.246&	0.337&	0.249&	0.344&	0.269&	0.369&	0.209&	0.301\\ 
    &720	&\textbf{\textcolor{red}{0.188}}	&\textbf{\textcolor{red}{0.285}}&	\underline{\textcolor{blue}{0.201}}	&\underline{\textcolor{blue}{0.294}}	&0.212	&0.299&	0.211&	0.306&	 0.218&	0.305&	0.225&	0.317&	0.280&	0.363&	0.284&	0.373&	0.299	&0.390&	0.245&	0.333\\ 
    \cmidrule(lr){2-24}  & Avg	&\textbf{\textcolor{red}{0.158}}&	\textbf{\textcolor{red}{0.254}}&	\underline{\textcolor{blue}{0.165}}&	\underline{\textcolor{blue}{0.259}}&	0.168&	\underline{\textcolor{blue}{0.259}}&	0.171&	0.270&	0.270&	0.304&	0.365	&0.174&	0.244	&0.334	&0.252	&0.344	&0.268	&0.365&	0.212&	0.300\\ 
    \midrule\multirow{5}{*}{\rotatebox{90}{Trafﬁc}}
    & 96&	0.430& \textbf{\textcolor{red}{0.249}}	&	0.413	& \underline{\textcolor{blue}{0.261}}	&0.458&	0.296	&0.428&	\underline{\textcolor{blue}{0.271}}&	\textbf{\textcolor{red}{0.376}}&0.268&	\underline{\textcolor{blue}{0.395}}&	0.268&	0.522&	0.290&	0.805&	0.493&	0.788&	0.499&	0.650&	0.396\\ 
    &192&0.455	&\textbf{\textcolor{red}{0.264}}	&	0.432&	0.271&	0.457&	0.294&	0.448&	0.282&	\textbf{\textcolor{red}{0.398}}&	0.276&	\underline{\textcolor{blue}{0.417}}&	0.276&	0.530&	0.293&	0.756&	0.474&	0.789&	0.505&	0.598&	0.370\\ 
    & 336& 	0.481& \underline{\textcolor{blue}{0.281}}&	0.450&	\textbf{\textcolor{red}{0.277}}&	0.470&	0.299&	0.473&	0.289&	\textbf{\textcolor{red}{0.415}}&0.283&	\underline{\textcolor{blue}{0.433}}&	{0.283}&	0.558&	0.305&	0.762&	0.477&	0.797&	0.508&	0.605&	0.373\\ 
    & 720	&0.509	&\textbf{\textcolor{red}{0.286}}&	0.486&0.295&	0.502&	0.314&	0.516&	0.307&	\textbf{\textcolor{red}{0.447}} & 0.287&	\underline{\textcolor{blue}{0.467}}&	\underline{\textcolor{blue}{0.302}}&	0.559&	0.328&	0.719	&0.449&	0.841	&0.523&	0.645&	0.394\\ 
  \cmidrule(lr){2-24}
& Avg
& 0.469
& \textbf{\textcolor{red}{0.270}}
& 0.445
& \underline{\textcolor{blue}{0.276}}
& 0.472
& 0.301
& 0.466
& 0.287
& \textbf{\textcolor{red}{0.409}}
& 0.279
& \underline{\textcolor{blue}{0.428}}
& 0.282
& 0.542
& 0.304
& 0.761
& 0.473
& 0.804
& 0.509
& 0.625
& 0.383
\\
    \midrule\multirow{5}{*}{\rotatebox{90}{Weather}}
    & 96&	\textbf{\textcolor{red}{0.155}}&	\textbf{\textcolor{red}{0.200}}&	\underline{\textcolor{blue}{0.157}}	&\textbf{\textcolor{red}{0.200}}&	0.158&	\underline{\textcolor{blue}{0.203}}&	\underline{\textcolor{blue}{0.157}}&	0.205&	0.166&	0.214&	0.174&	0.214&	0.158&	0.230&	0.202&	0.261&	0.221	&0.306	&0.196	&0.255\\ 
    & 192&	\textbf{\textcolor{red}{0.203}}&	\textbf{\textcolor{red}{0.244}}&	0.206&	\underline{\textcolor{blue}{0.245}}&	0.207	&0.247&	\underline{\textcolor{blue}{0.204}}&	0.247&	0.217&	0.254&	0.221&	0.254&	0.206&	0.277&	0.242&	0.298&	0.261&	0.340&	0.237&	0.296\\ 
    & 336&	0.265&	\underline{\textcolor{blue}{0.289}}	&\underline{\textcolor{blue}{0.262}}&	\textbf{\textcolor{red}{0.287}}&	\underline{\textcolor{blue}{0.262}}&	\underline{\textcolor{blue}{0.289}}&	\textbf{\textcolor{red}{0.261}}&	0.290&	0.282&	0.296&	0.278&	0.296&	0.272&	0.335&	0.287&	0.335&	0.309&	0.378&	0.283&	0.335\\ 
    & 720&	\underline{\textcolor{blue}{0.344}}&	\underline{\textcolor{blue}{0.342}}&	\underline{\textcolor{blue}{0.344}}&	\underline{\textcolor{blue}{0.342}}&	\underline{\textcolor{blue}{0.344}}&	0.344&	\textbf{\textcolor{red}{0.340}}&	\textbf{\textcolor{red}{0.341}}&	0.356&	0.347&	0.358&	0.347&	0.398&	0.418&	0.351&	0.386&	0.377&	0.427&	0.345&	0.381\\ 
    \cmidrule(lr){2-24}  & Avg	&\underline{\textcolor{blue}{0.242}}&	\textbf{\textcolor{red}{0.269}}&	\underline{\textcolor{blue}{0.242}}&	\textbf{\textcolor{red}{0.269}}&	0.243&	\underline{\textcolor{blue}{0.271}}&	\textbf{\textcolor{red}{0.241}}&	\underline{\textcolor{blue}{0.271}}&	0.255&	0.278&	0.258&	0.278&	0.259&	0.315&	0.271&	0.320&	0.292&	0.363&	0.265&	0.317\\
    \midrule\multirow{5}{*}{\rotatebox{90}{Solar}}
    & 96&	\underline{\textcolor{blue}{0.177}}&	\textbf{\textcolor{red}{0.231}}&	\textbf{\textcolor{red}{0.173}}&	\underline{\textcolor{blue}{0.233}}&	0.190&	0.247&	0.215&	0.295&	0.230&	0.246&	0.203&	0.237&	0.310&	0.331&	0.312&	0.399&	0.237&	0.344&	0.290&	0.378\\
    & 192	&\textbf{\textcolor{red}{0.198}}&	\textbf{\textcolor{red}{0.250}}&	\underline{\textcolor{blue}{0.199}}&	\underline{\textcolor{blue}{0.257}}&	0.210&	0.266&	0.236&	0.301&	0.253&	0.267&	0.233&	0.261&	0.734&	0.725&	0.339&	0.416&	0.280&	0.380&	0.320&	0.398\\
    & 336&	\textbf{\textcolor{red}{0.208}}&	\textbf{\textcolor{red}{0.260}}&	\underline{\textcolor{blue}{0.211}}&	\underline{\textcolor{blue}{0.263}}&	0.217&	0.266&	0.252&	0.307&	0.269&	0.276&	0.248&	0.273&	0.750&	0.735&	0.368&	0.430&	0.304&	0.389&	0.353&	0.415\\
    & 720&	\textbf{\textcolor{red}{0.209}}&	\textbf{\textcolor{red}{0.256}}&	\textbf{\textcolor{red}{0.209}}&	0.270&	\underline{\textcolor{blue}{0.223}}&	\underline{\textcolor{blue}{0.266}}&	0.244&	0.305&	0.272&	0.275&	0.249&	0.275&	0.769&	0.765&	0.370&	0.425&	0.308&	0.388&	0.356&	0.413\\ 
    \cmidrule(lr){2-24}  & Avg&	\textbf{\textcolor{red}{0.198}}&	\textbf{\textcolor{red}{0.249}}&	\textbf{\textcolor{red}{0.198}}&	\underline{\textcolor{blue}{0.256}}&	\underline{\textcolor{blue}{0.210}}&	0.261&	0.237&	0.302&	0.256&	0.266&	0.233&	0.262&	0.641&	0.639&	0.347&	0.418&	0.282&	0.375&	0.330&	0.401\\
    \midrule\multirow{5}{*}{\rotatebox{90}{PEMS03}}
    &12	&\textbf{\textcolor{red}{0.059}}&	\textbf{\textcolor{red}{0.155}}&	\underline{\textcolor{blue}{0.060}}&	\underline{\textcolor{blue}{0.161}}&	0.066&	0.172&	0.070&	0.173&	0.064&	0.165&	0.071&	0.174&	0.090&	0.203&	0.178&	0.305&	0.066&	0.172&	0.122&	0.243\\ 
    & 24&	\textbf{\textcolor{red}{0.070}}&	\textbf{\textcolor{red}{0.170}}&	\underline{\textcolor{blue}{0.077}}&	\underline{\textcolor{blue}{0.182}}&	0.089&	0.201&	0.092&	0.194&	0.083	&0.188&	0.093	&0.201&	0.121&	0.240&	0.257	&0.371&	0.085	&0.198&	0.201&	0.317\\
    & 48&	\textbf{\textcolor{red}{0.095}}&	\textbf{\textcolor{red}{0.195}}	&\underline{\textcolor{blue}{0.104}}&	\underline{\textcolor{blue}{0.215}}&	0.136&	0.247&	0.129&	0.229&	0.114&	0.223&	0.125&	0.236&	0.202&	0.317&	0.379&	0.463&	0.127&	0.238&	0.333&	0.425\\
    & 96&	\textbf{\textcolor{red}{0.145}}&	\textbf{\textcolor{red}{0.248}}&	\underline{\textcolor{blue}{0.148}}	&\underline{\textcolor{blue}{0.253}}&	0.182&	0.282&	0.157&	0.261&	0.156&	0.264&	0.164&	0.275&	0.262&	0.367&	0.490&	0.539&	0.178&	0.287&	0.457&	0.515\\
    \cmidrule(lr){2-24} & Avg&	\textbf{\textcolor{red}{0.092}}&	\textbf{\textcolor{red}{0.192}}&	\underline{\textcolor{blue}{0.097}}&	\underline{\textcolor{blue}{0.203}}&	0.118&	0.226&	0.112&	0.214&	0.104&	0.210&	0.113&	0.222&	0.169&	0.282&	0.326&	0.420&	0.114&	0.224&	0.278&	0.375\\
    \midrule\multirow{5}{*}{\rotatebox{90}{PEMS04}}
    & 12	&\textbf{\textcolor{red}{0.064}}&	\textbf{\textcolor{red}{0.161}}	&\underline{\textcolor{blue}{0.067}}&	\underline{\textcolor{blue}{0.166}}&	0.078&	0.186&	0.074&	0.178&	0.074&	0.176&	0.078&	0.183&	0.098&	0.218&	0.219&	0.340&	0.073&	0.177&	0.148&	0.272\\
    & 24&	\textbf{\textcolor{red}{0.071}}&	\textbf{\textcolor{red}{0.170}}&	\underline{\textcolor{blue}{0.077}}&	\underline{\textcolor{blue}{0.181}}&	0.099&	0.212&	0.087&	0.195&	0.088&	0.194&	0.095&	0.205&	0.131&	0.256&	0.292&	0.398&	0.084&	0.193& 0.224&	0.340\\
    & 48&	\textbf{\textcolor{red}{0.083}}&	\textbf{\textcolor{red}{0.185}}&	\underline{\textcolor{blue}{0.097}}&	\underline{\textcolor{blue}{0.206}}&	0.133&	0.248&	0.110&	0.214&	0.110&	0.219&	0.120&	0.233&	0.205&	0.326&	0.409&	0.478	&0.099	&0.211&	0.355&	0.437\\
    & 96&	\textbf{\textcolor{red}{0.099}}&	\textbf{\textcolor{red}{0.203}}&	\underline{\textcolor{blue}{0.123}}&	\underline{\textcolor{blue}{0.233}}&	0.167&	0.281&	0.148&	0.251&	0.135&	0.244&	0.150&	0.262& 0.402&	0.457&	0.492&	0.532&	0.114&	0.227&	0.452&	0.504\\
    \cmidrule(lr){2-24} & Avg&	\textbf{\textcolor{red}{0.079}}&	\textbf{\textcolor{red}{0.180}}&	\underline{\textcolor{blue}{0.091}}&	\underline{\textcolor{blue}{0.197}}&	0.119&	0.232&	0.105&	0.210&	0.102&	0.209&	0.111&	0.221&	0.209&	0.314&	0.353&	0.437&	0.093&	0.202&	0.295&	0.388\\
    \midrule\multirow{5}{*}{\rotatebox{90}{PEMS07}}
    &12&	\textbf{\textcolor{red}{0.050}}&	\textbf{\textcolor{red}{0.141}}&	\underline{\textcolor{blue}{0.051}}&	\underline{\textcolor{blue}{0.143}}&	0.062&	0.162&	0.057&	0.152&	0.057&	0.152&	0.067&	0.165&	0.094&	0.200&	0.173&	0.304&	0.068&	0.171&	0.115&	0.242\\
    & 24&	\textbf{\textcolor{red}{0.060}}&	\textbf{\textcolor{red}{0.153}}&	\underline{\textcolor{blue}{0.063}}&	\underline{\textcolor{blue}{0.159}}&	0.086&	0.192&	0.079&	0.179&	0.073&	0.173&	0.088&	0.190&	0.139&	0.247&	0.271&	0.383&	0.119&	0.225&	0.210&	0.329\\
    & 48&	\textbf{\textcolor{red}{0.076}}&	\textbf{\textcolor{red}{0.170}}&	\underline{\textcolor{blue}{0.081}}&	\underline{\textcolor{blue}{0.179}}&	0.128&	0.234&	0.099&	0.191&	0.096&	0.195&	0.110&	0.215&	0.311&	0.369&	0.446&	0.495&	0.149&	0.237&	0.398&	0.458\\
    & 96&	\textbf{\textcolor{red}{0.094}}&	\textbf{\textcolor{red}{0.188}}&	\underline{\textcolor{blue}{0.103}}&	\underline{\textcolor{blue}{0.203}}&	0.176&	0.268&	0.107&	0.205&	0.120&	0.218&	0.139&	0.245&	0.396&	0.442&	0.628&	0.577&	0.141&	0.234&	0.594&	0.553\\
    \cmidrule(lr){2-24} & Avg&	\textbf{\textcolor{red}{0.070}}&	\textbf{\textcolor{red}{0.163}}&	\underline{\textcolor{blue}{0.074}}&	\underline{\textcolor{blue}{0.171}}&	0.113&	0.214&	0.086&	0.182&	0.087&	0.189&	0.101&	0.204&	0.235&	0.315&	0.380&	0.440&	0.119&	0.217&	0.329&	0.396\\
    \midrule\multirow{5}{*}{\rotatebox{90}{PEMS08}}
    & 12&	\textbf{\textcolor{red}{0.069}}&	\textbf{\textcolor{red}{0.164}}&	\underline{\textcolor{blue}{0.071}}&	\underline{\textcolor{blue}{0.170}}&	0.082&	0.185&	0.075&	0.176&	0.074&	0.171&	0.079&	0.182&	0.165&	0.214&	0.227&	0.343&	0.087&	0.184&	0.154&	0.276\\
    & 24&	\textbf{\textcolor{red}{0.087}}&	\textbf{\textcolor{red}{0.185}}&	\underline{\textcolor{blue}{0.096}}&	\underline{\textcolor{blue}{0.196}}&	0.117&	0.226&	0.102&	0.201&	0.104&	0.201&	0.115&	0.219&	0.215&	0.260&	0.318&	0.409&	0.122&	0.221&	0.248&	0.353\\
    & 48&	\textbf{\textcolor{red}{0.125}}&	\textbf{\textcolor{red}{0.212}}&	\underline{\textcolor{blue}{0.149}}&	0.244&	0.169&	0.268&	0.158&	0.248&	0.164&	0.253&	0.186&	\underline{\textcolor{blue}{0.235}}&	0.315&	0.355&	0.497	&0.510&	0.189&	0.270	&0.440&	0.470\\
    & 96&	\textbf{\textcolor{red}{0.204}}&	\textbf{\textcolor{red}{0.250}}&	0.253&	0.309&	0.233&	0.306&	0.366&	0.377&	\underline{\textcolor{blue}{0.211}}&	\underline{\textcolor{blue}{0.253}}&	0.221&	0.267&	0.377&	0.397&	0.721&	0.592&	0.236&	0.300&	0.674&	0.565\\
    \cmidrule(lr){2-24} & Avg&	\textbf{\textcolor{red}{0.121}}&	\textbf{\textcolor{red}{0.203}}&	0.142&	0.230&	0.150&	0.246&	0.175&	0.251&	0.138&	0.219&	0.150&	0.226&	0.268&	0.307&	0.441&	0.464&	0.159&	0.244&	0.379&	0.416 \\
    
    \midrule
    \multicolumn{2}{c|}{{{$1^{\text{st}}$ Count}}}       & \textbf{\textcolor{red}{38}}& 	\textbf{\textcolor{red}{42}}& 	3& 	\underline{\textcolor{blue}{8}}& 	6& 	7& 	\underline{\textcolor{blue}{8}}& 	7& 	6& 	0& 	0& 	0& 	0	& 0	& 0& 	0& 	0	& 0	& 0& 	0 \\
\bottomrule
    \end{tabular}}
    }
\end{table}

\noindent \textbf{Datasets} To evaluate the effectiveness of the proposed architecture, we conduct experiments on 6 widely used real-world multivariate time series forecasting benchmarks, including ETT (with 4 subsets)~\cite{informer}, Weather, Solar~\cite{lstnet}, Electricity (ECL), Traffic~\cite{autoformer}, and PEMS (with 4 subsets)~\cite{scinet}. These datasets span diverse application domains and are commonly used to benchmark long-term multivariate time-series forecasting performance. Table~\ref{tab:dataset_stats} provides an overview of the datasets, including their channels, sampling intervals, and total timesteps. Detailed descriptions of the experimental datasets, including the training, validation, and test splits, are provided in Appendix \ref{app:det_data}.

\noindent \textbf{Baselines} We compare CARNet with representative models from periodic modeling, transformer-based forecasting, and efficient non-transformer approaches. Specifically, the baselines include recent cycle-aware methods CycleNet~\cite{cyclenet} and TQNet~\cite{tqnet}; transformer-based models iTransformer~\cite{itransformer}, Crossformer~\cite{crossformer}, and TimeXer~\cite{timexer}; and efficient non-transformer methods SOFTS~\cite{softs}, TiDE~\cite{tide}, SCINet~\cite{scinet}, and DLinear~\cite{dlinear}. These methods constitute strong and widely used baselines for multivariate time series forecasting.

\subsection{Main Results}

\noindent The comparative results on 12 real-world multivariate forecasting datasets are summarized in Table~\ref{tab:detailed_res}. Lower MSE and MAE values indicate better forecasting accuracy. Overall, CARNet achieves the best performance in 38 out of 48 settings in terms of MSE and 42 out of 48 settings in terms of MAE across all datasets and forecasting horizons. When averaged across forecasting horizons, CARNet attains the lowest error on 8 out of 12 datasets in MSE and 10 out of 12 datasets in MAE, demonstrating strong and consistent performance.

CARNet consistently outperforms recent periodic modeling approaches such as TQNet~\cite{tqnet} and CycleNet~\cite{cyclenet}, as well as attention-based methods, including TimeXer~\cite{timexer} and attention-free baselines such as SOFTS~\cite{softs}. Notably, CARNet shows clear advantages on high-dimensional multivariate datasets such as Electricity and on short-term traffic forecasting benchmarks in the PEMS family, highlighting its effectiveness in modeling complex cross-variate dependencies under structured temporal dynamics.
These results suggest that integrating cycle-conditioned representations with structured core aggregation enables CARNet to effectively capture global temporal structure while maintaining efficient cross-variate interaction modeling.

% \begin{figure*}[t]
% \centering
% \includegraphics[width=\textwidth]{Figures/Embedding.pdf}
% \caption{Visualization of embeddings using t-SNE.\mk{remove this plot for now}}
% \label{Embedding visualize}
% \end{figure*}

\subsection{Ablation Study}
\noindent \textbf{Component Analysis.} 
We conduct a component-level ablation study to quantify the contribution of each module in CARNet. Specifically, we construct several variants by removing or modifying key components, including cycle conditioning, Multihead Core Aggregation (MHCA), and core redistribution, to isolate their roles in cross-variate dependency modeling. All variants are evaluated on six representative multivariate time series datasets under identical training settings (optimization strategy, learning rate schedule, and training epochs), ensuring that performance differences arise solely from architectural changes.

\begin{table}[!h]
\centering
\caption{Averaged results of component-level ablation on six datasets.
wo\_MHCA, wo\_CCCA, and wo\_CCCR denote model variants where Multihead Core Aggregation, Cycle-Conditioned Core Aggregation, and Cycle-Conditioned Core Redistribution are removed, respectively.}
\label{tab:comp_avg}
\normalfont
\setlength{\tabcolsep}{7pt}
\renewcommand{\arraystretch}{0.9}
\resizebox{\columnwidth}{!}{
\tiny
\begin{tabular}{c|cc|cc|cc|cc}
\toprule
\textbf{Dataset} &
\multicolumn{2}{c|}{CARNet (Ours)} &
\multicolumn{2}{c|}{wo\_MHCA} &
\multicolumn{2}{c|}{wo\_CCCA} &
\multicolumn{2}{c}{wo\_CCCR} \\
\cmidrule(lr){2-3}\cmidrule(lr){4-5}\cmidrule(lr){6-7}\cmidrule(lr){8-9}
\textbf{Metric} & MSE & MAE & MSE & MAE & MSE & MAE & MSE & MAE \\
\midrule
ETTm1   &  \textbf{\textcolor{red}{0.375}} & \textbf{\textcolor{red}{0.392}}
& 0.380 & 0.395
& 0.379 & 0.396
& 0.384 & 0.395 \\
Solar   & \textbf{\textcolor{red}{0.198}} & \textbf{\textcolor{red}{0.249}}
& 0.203 & 0.260
& 0.209 & 0.265
& 0.211 & 0.259 \\
Weather & \textbf{\textcolor{red}{0.242}} & \textbf{\textcolor{red}{0.269}}
& 0.243 & 0.270
& 0.243 & 0.270
& 0.253 & 0.273 \\
PEMS04
& \textbf{\textcolor{red}{0.079}} & \textbf{\textcolor{red}{0.180}}
& 0.083 & 0.182
& 0.082 & 0.182
& 0.103 & 0.209 \\

PEMS08  & \textbf{\textcolor{red}{0.121}} & \textbf{\textcolor{red}{0.203}}
& 0.130 & 0.207
& 0.125 & 0.206
& 0.153 & 0.237 \\
ECL     & \textbf{\textcolor{red}{0.158}} & \textbf{\textcolor{red}{0.254}}
& 0.162 & 0.256
& 0.160 & 0.255
& 0.173 & 0.264 \\
\bottomrule
\end{tabular}
}
\end{table}

As shown in Table~\ref{tab:comp_avg}, removing any individual component consistently degrades forecasting performance, confirming that each module contributes to the overall effectiveness of CARNet. Notably, eliminating Cycle-Conditioned Core Redistribution results in the largest performance drop, highlighting the critical role of integrating periodic information during the redistribution stage. These findings demonstrate that both cycle conditioning and MHCA are essential for robust cross-variate interaction modeling. Full result is provided in Appendix \ref{app:full component}.

\begin{figure}[!h]
    \centering
    \includegraphics[width=\columnwidth]{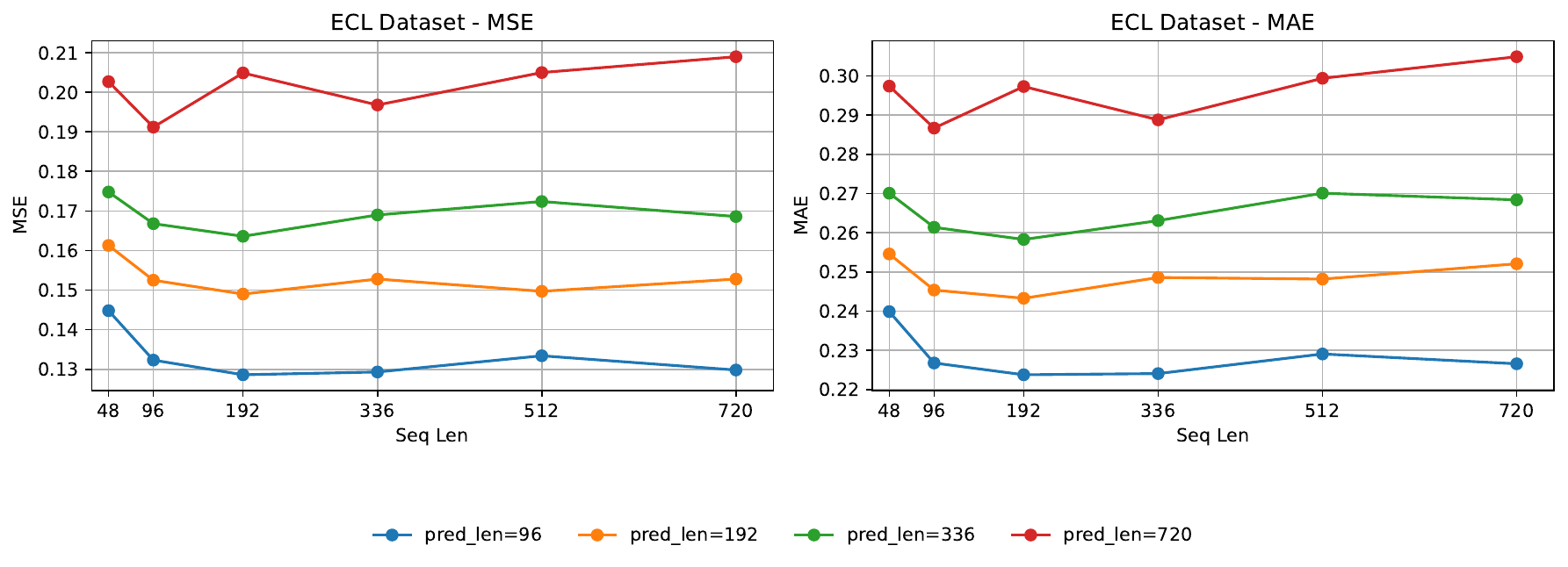}
    \caption{Performance on different look-back length on ECL dataset.}
    \label{ECL_lookback}
\end{figure}

\noindent \textbf{Look-Back Length Ablation.} Intuitively, longer look-back windows provide more historical context, but excessively long inputs may introduce redundant or noisy information that hinders forecasting. As shown in Figure~\ref{ECL_lookback}, CARNet remains stable across a wide range of look-back lengths and prediction horizons on the ECL dataset. Performance generally improves as the input length increases from 48 to 192, indicating that CARNet effectively utilizes additional historical information to capture periodic and cross-variate patterns. Beyond this range, the gains become marginal or slightly decline, suggesting diminishing returns from overly long input windows. Overall, these results demonstrate the robustness of CARNet to input window variations.

\noindent \textbf{Cycle-Conditioned Integration vs. Residual Cycle Decomposition.}
To examine whether CARNet’s gains come only from recurrent cycle modeling, we perform an ablation by integrating Residual Cycle Forecasting (RCF)~\cite{cyclenet} into the vanilla SOFTS framework. Following this setup, SOFTS is applied to the residual series after removing the recurrent cycle component, and the cycle is added back to the final prediction. We evaluate this RCF+SOFTS variant on six benchmark datasets across four prediction horizons using the same training settings as the main experiments.

\begin{figure}[!h]
\centering
\includegraphics[width=0.7\textwidth]{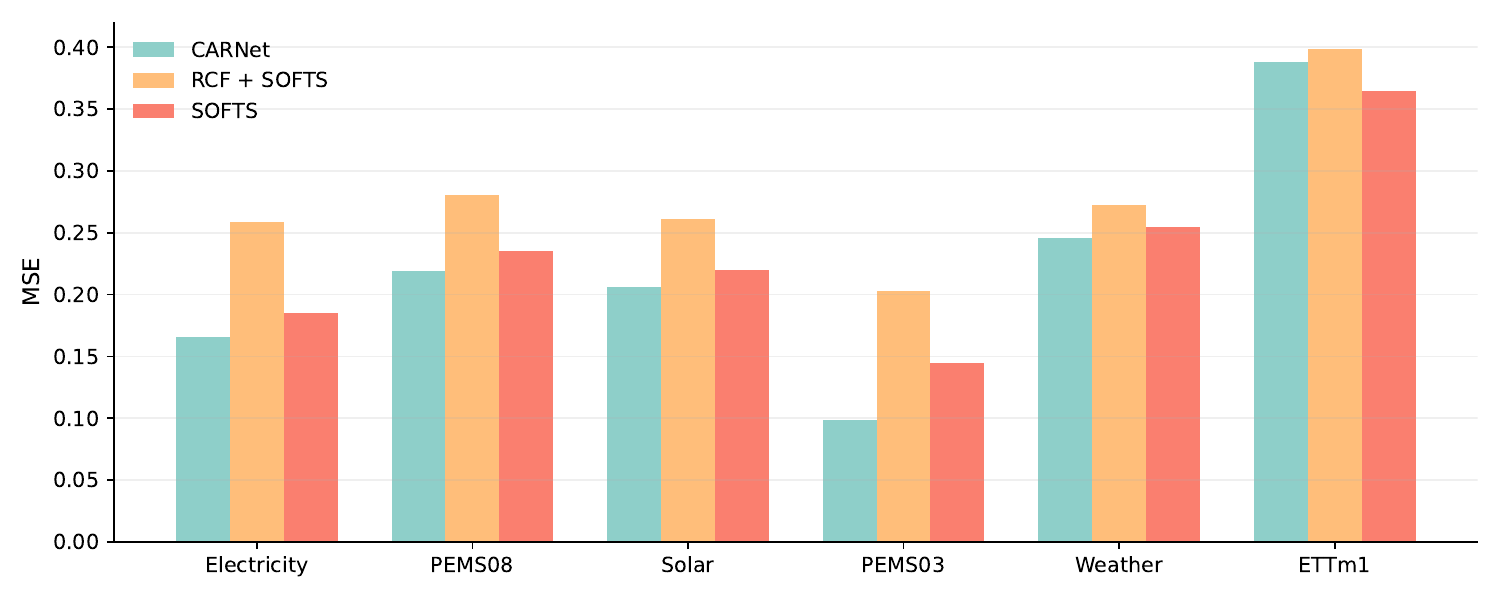}
\caption{Comparison between CARNet and SOFTS augmented with Residual Cycle Forecasting (RCF) across six datasets, averaged over four prediction horizons.}
\label{cycle integration}
\end{figure}
As shown in Figure~\ref{cycle integration}, CARNet consistently outperforms RCF+SOFTS, indicating that its Cycle-Conditioned Core Aggregation and redistribution mechanisms integrate periodic information more effectively than simple residual-based cycle decomposition.

\begin{figure}[!h]
\centering
\includegraphics[width=0.7\columnwidth]{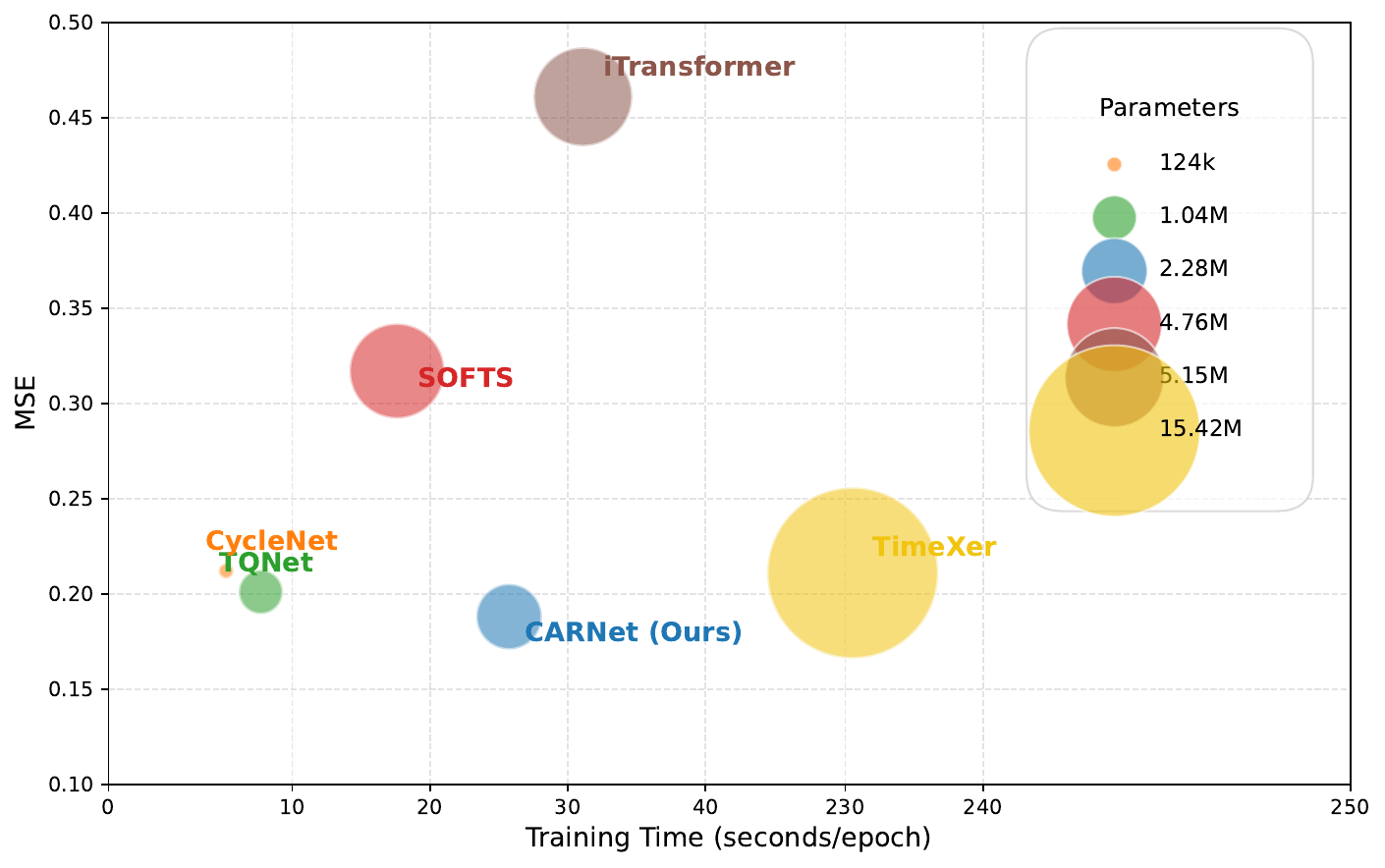}
\caption{Efficiency Analysis of CARNet On ECL Dataset.} 
\label{Efficiency}
\end{figure}

\noindent \textbf{Efficiency Analysis.}  We analyze the efficiency of CARNet in terms of parameter count and training speed on ECL dataset, as shown in Figure~\ref{Efficiency}. All models are re-run on our hardware using their officially reported hyperparameters to ensure a fair comparison. Building on the linear-complexity, core-based interaction paradigm, CARNet achieves stronger forecasting performance than existing state-of-the-art methods by introducing cycle-conditioned core aggregation and redistribution together with the proposed Multi-Head Core Aggregation mechanism.

On the ECL dataset, CARNet uses fewer parameters than SOFTS, which can be attributed to its ability to achieve improved performance with slightly smaller hidden dimensions under comparable settings. Compared to attention-based models such as TQNet, CARNet has a higher parameter count and longer per-epoch training time. This difference stems from architectural choices rather than interaction complexity: TQNet employs a single-stack design with a very small embedding dimension, whereas CARNet, similar to SOFTS, benefits from multiple encoder layers to better capture complex cross-variate dependencies on datasets such as ECL, PEMS, and Traffic. Importantly, although TQNet relies on attention mechanisms, its practical efficiency is achieved through shallow depth and reduced embedding size. In contrast, CARNet preserves linear-complexity interaction modeling while scaling depth to improve representation quality and leading to a favorable accuracy–efficiency trade-off across diverse benchmarks.

\begin{figure}[!h]
\centering
\includegraphics[width=0.7\columnwidth]{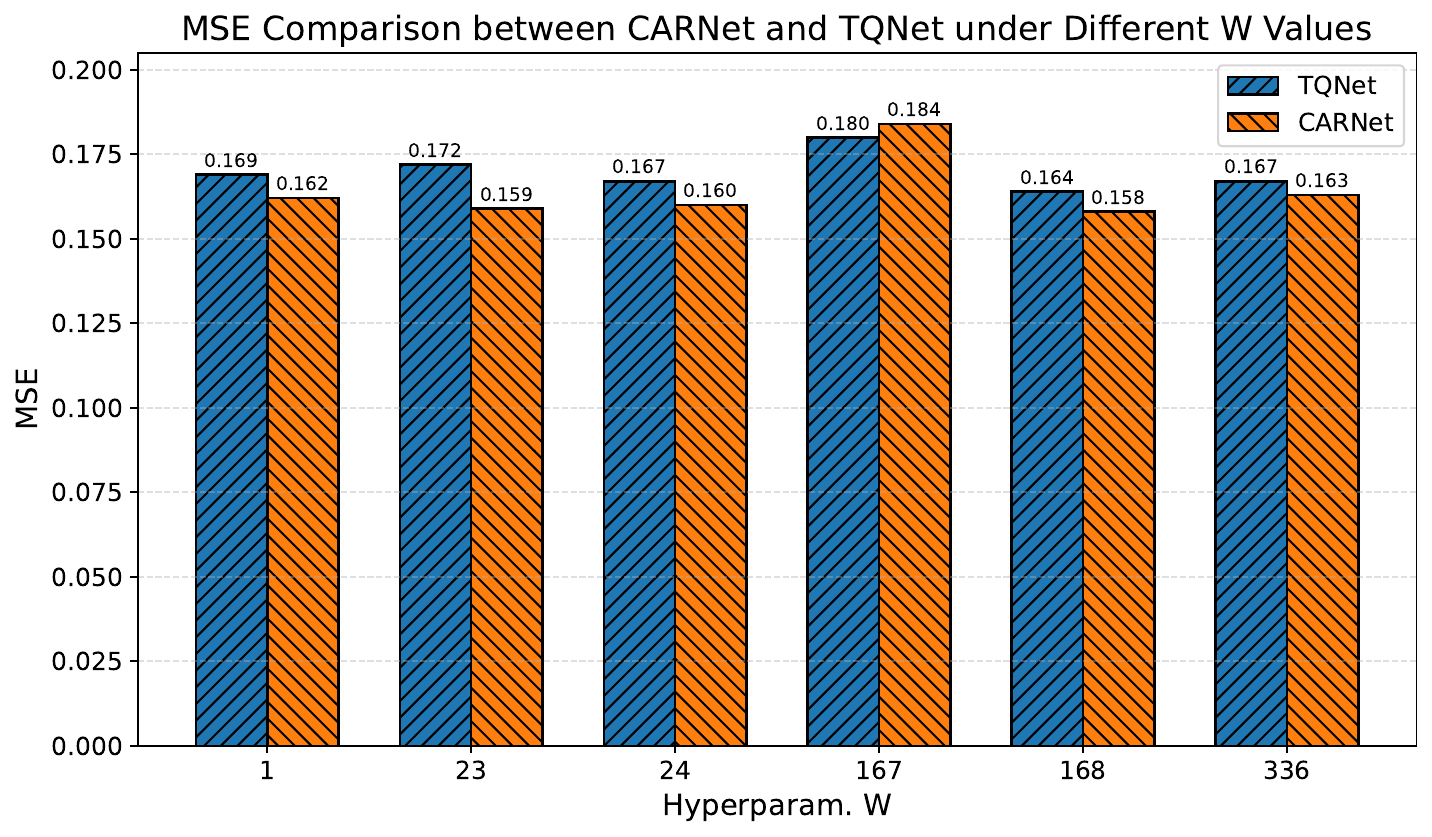}
\caption{Effect of cycle length $W$ on forecasting performance (averaged MSE and MAE over four prediction horizons) on the ECL dataset. CARNet is compared against TQNet~\cite{tqnet} across the same set of cycle lengths, with $W{=}168$ corresponding to the dominant weekly periodicity.}
\label{Cycle Length}
\end{figure}

\noindent \textbf{Cycle Length Ablation.} To assess the sensitivity of CARNet to the choice of cycle length $W$, we evaluate the model across multiple values of $W$ on the ECL dataset, averaged over four prediction horizons, while keeping all other hyperparameters fixed. The cycle lengths considered follow those reported in TQNet~\cite{tqnet}, enabling a direct comparison. As shown in Figure~\ref{Cycle Length}, both CARNet and TQNet exhibit performance degradation when $W$ deviates from the stable value of $W{=}168$, which corresponds to the dominant weekly periodicity in ECL. However, the degradation of CARNet is considerably smaller across all tested values.

Notably, CARNet achieves performance at $W{=}24$, representing daily periodicity and a harmonic of the weekly cycle that remains close to its performance at $W{=}168$, whereas TQNet degrades more sharply under the same condition. Furthermore, the average MSE across $W{=}23$ and $W{=}24$ remains substantially more stable for CARNet than for TQNet, suggesting that CARNet is better able to exploit harmonically related periodicities. These results indicate that the Cycle-Conditioned Core Aggregation and redistribution mechanisms confer greater robustness to cycle length misspecification, reflecting a more principled integration of periodic structure into the cross-variate interaction framework.

\noindent \textbf{Evaluation Under Fixed Training Epoch.} To ensure a fair comparison with recent state-of-the-art models such as TimeXer~\cite{timexer} and iTransformer~\cite{itransformer}, which are trained for only 10 epochs, we evaluate CARNet under the same training budget. CARNet is trained for 10 epochs while keeping all other hyperparameters identical to the full-training setup. We also retrain TQNet~\cite{tqnet} and SOFTS~\cite{softs} under the same setting to eliminate any advantage arising from longer optimization. Experiments are conducted on 5 representative datasets across 4 forecasting horizons, covering diverse temporal characteristics including electricity consumption, traffic flow, and weather. As shown in Table~\ref{tab:same epoch}, CARNet achieves the highest number of best results in both MSE and MAE, indicating that the proposed Cycle-Conditioned Core Aggregation and redistribution framework converges effectively even under a limited training budget. This suggests that the integration of periodic structure into the core interaction mechanism provides a strong inductive bias that accelerates learning and reduces dependence on extended training.

\begin{table}[!h]
\caption{Performance comparison on five datasets under a fixed training budget of 10 epochs. Results for TimeXer \cite{timexer} and iTransformer \cite{itransformer} are reported from TQNet \cite{tqnet}.}
\label{tab:same epoch}
\scriptsize
\setlength{\tabcolsep}{4.5pt}
\renewcommand{\arraystretch}{0.7}

\resizebox{\textwidth}{!}{
\begin{tabular}{c c | cc | cc | cc | cc | cc}
\toprule
Dataset & Pred
& \multicolumn{2}{c}{CARNet (Ours)}
& \multicolumn{2}{c}{TQNet}
& \multicolumn{2}{c}{SOFTS}
& \multicolumn{2}{c}{TimeXer}
& \multicolumn{2}{c}{iTransformer} \\
\cmidrule(lr){3-4}\cmidrule(lr){5-6}\cmidrule(lr){7-8}\cmidrule(lr){9-10}\cmidrule(lr){11-12}
& & MSE & MAE & MSE & MAE & MSE & MAE & MSE & MAE & MSE & MAE \\
\midrule

\multirow{5}{*}{ETTm1}
& 96  & \textbf{\textcolor{red}{0.310}} & \textbf{\textcolor{red}{0.353}} & 0.311 & \textbf{\textcolor{red}{0.353}} & 0.328 & 0.365 & 0.318 & 0.356 & 0.334 & 0.368 \\
& 192 & \textbf{\textcolor{red}{0.355}} & \textbf{\textcolor{red}{0.379}} & 0.357 & 0.379 & 0.382 & 0.393 & 0.362 & 0.383 & 0.377 & 0.391 \\
& 336 & 0.391 & 0.405 & \textbf{\textcolor{red}{0.388}} & \textbf{\textcolor{red}{0.400}} & 0.419 & 0.415 & 0.395 & 0.407 & 0.426 & 0.420 \\
& 720 & 0.454 & \textbf{\textcolor{red}{0.440}} & \textbf{\textcolor{red}{0.451}} & 0.440 & 0.489 & 0.458 & 0.452 & 0.441 & 0.491 & 0.459 \\
\midrule
& Avg
& 0.378& \textbf{\textcolor{red}{0.393}}
& \textbf{\textcolor{red}{0.377}} & \textbf{\textcolor{red}{0.393}}
& 0.405 & 0.408
& 0.382 & 0.397
& 0.407 & 0.410 \\
\midrule

\multirow{5}{*}{ECL}
& 96  & \textbf{\textcolor{red}{0.132}} & \textbf{\textcolor{red}{0.229}} & 0.138 & 0.234 & 0.145 & 0.236 & 0.140 & 0.242 & 0.148 & 0.240 \\
& 192 & \textbf{\textcolor{red}{0.151}} & \textbf{\textcolor{red}{0.245}} & 0.156 & 0.250 & 0.159 & 0.249 & 0.157 & 0.256 & 0.162 & 0.253 \\
& 336 & \textbf{\textcolor{red}{0.165}} & \textbf{\textcolor{red}{0.262}} & 0.172 & 0.266 & 0.180 & 0.271 & 0.176 & 0.275 & 0.178 & 0.269 \\
& 720 & \textbf{\textcolor{red}{0.192}} & \textbf{\textcolor{red}{0.288}} & 0.203 & 0.294 & 0.210 & 0.297 & 0.211 & 0.306 & 0.225 & 0.317 \\
\midrule
& Avg
& \textbf{\textcolor{red}{0.160}} & \textbf{\textcolor{red}{0.256}}
& 0.167 & 0.261
& 0.174 & 0.263
& 0.171 & 0.270
& 0.178 & 0.270 \\
\midrule

\multirow{5}{*}{Weather}
& 96  & \textbf{\textcolor{red}{0.156}} & \textbf{\textcolor{red}{0.201}} & 0.158 & 0.202 & 0.173 & 0.215 & 0.157 & 0.205 & 0.166 & 0.214 \\
& 192 & \textbf{\textcolor{red}{0.203}} & \textbf{\textcolor{red}{0.244}} & 0.207 & 0.247 & 0.218 & 0.256 & 0.204 & 0.247 & 0.221 & 0.254 \\
& 336 & 0.266 & 0.290 & 0.263 & \textbf{\textcolor{red}{0.287}} & 0.280 & 0.300 & \textbf{\textcolor{red}{0.261}} & 0.290 & 0.278 & 0.296 \\
& 720 & 0.345 & 0.342 & 0.344 & 0.342 & 0.359 & 0.351 & \textbf{\textcolor{red}{0.340}} & \textbf{\textcolor{red}{0.341}} & 0.358 & 0.347 \\
\midrule
& Avg
& 0.243 & \textbf{\textcolor{red}{0.269}}
& 0.243 & 0.270
& 0.258 & 0.281
& \textbf{\textcolor{red}{0.241}} & 0.271
& 0.256 & 0.278 \\
\midrule

\multirow{5}{*}{Solar}
& 96  & 0.178 & 0.231 & \textbf{\textcolor{red}{0.172}} & 0.234 & 0.201 & \textbf{\textcolor{red}{0.229}} & 0.215 & 0.295 & 0.203 & 0.237 \\
& 192 & 0.199 & \textbf{\textcolor{red}{0.250}} & \textbf{\textcolor{red}{0.198}} & 0.257 & 0.228 & 0.253 & 0.236 & 0.301 & 0.233 & 0.261 \\
& 336 & 0.209 & \textbf{\textcolor{red}{0.261}} & \textbf{\textcolor{red}{0.206}} & 0.264 & 0.242 & 0.269 & 0.252 & 0.307 & 0.248 & 0.273 \\
& 720 & \textbf{\textcolor{red}{0.210}} & \textbf{\textcolor{red}{0.256}} & \textbf{\textcolor{red}{0.210}} & 0.270 & 0.246 & 0.273 & 0.244 & 0.305 & 0.249 & 0.275 \\
\midrule
& Avg
& 0.199 & \textbf{\textcolor{red}{0.250}}
& \textbf{\textcolor{red}{0.197}} & 0.256
& 0.229 & 0.256
& 0.237 & 0.302
& 0.233 & 0.262 \\
\midrule

\multirow{5}{*}{PEMS08}
& 96  & \textbf{\textcolor{red}{0.070}} & \textbf{\textcolor{red}{0.169}} & 0.073 & 0.174 & 0.283 & 0.336 & 0.075 & 0.176 & 0.079 & 0.182 \\
& 192 & \textbf{\textcolor{red}{0.090}} & \textbf{\textcolor{red}{0.188}} & 0.101 & 0.204 & 0.369 & 0.360 & 0.102 & 0.201 & 0.115 & 0.219 \\
& 336 & \textbf{\textcolor{red}{0.134}} & \textbf{\textcolor{red}{0.223}} & 0.165 & 0.260 & 0.398 & 0.337 & 0.158 & 0.248 & 0.186 & 0.235 \\
& 720 & 0.227 & \textbf{\textcolor{red}{0.264}} & 0.298 & 0.339 & 0.474 & 0.388 & 0.366 & 0.377 & \textbf{\textcolor{red}{0.221}} & 0.267 \\
\midrule
& Avg
& \textbf{\textcolor{red}{0.130}} & \textbf{\textcolor{red}{0.211}}
& 0.159 & 0.244
& 0.381 & 0.355
& 0.175 & 0.251
& 0.150 & 0.226 \\
\bottomrule
\end{tabular}
}
\end{table}

\begin{figure*}[h!]
\centering
\includegraphics[width=1\columnwidth]{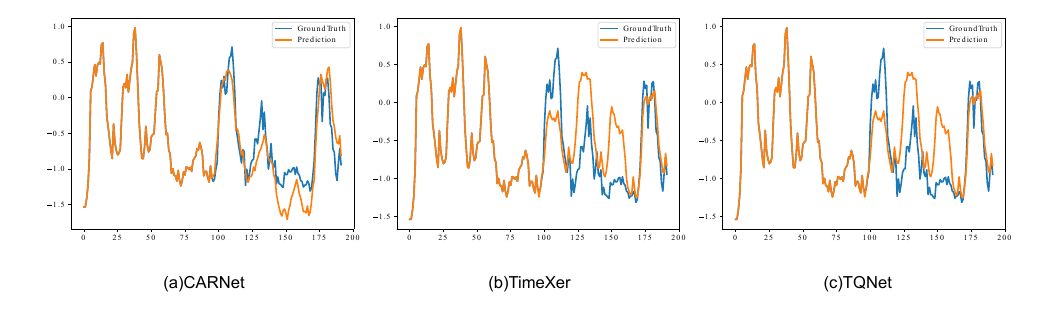}
\caption{Visualization of forecasting results on the ECL dataset with look-back window 96 and prediction horizon 96.}
\label{ECL prediction}
\end{figure*}

\noindent \textbf{Visualization on prediction.} To provide an intuitive understanding of the forecasting behavior of different models, we visualize the actual and predicted time series on the ECL dataset with both the look-back window and prediction horizon set to 96. We compare CARNet with two strong baselines, TQNet and TimeXer, to highlight differences in modeling periodic patterns and long-range dependencies. As shown in Figure~\ref{ECL prediction}, CARNet produces predictions that are more accurate with the ground-truth series, while TQNet and TimeXer exhibit noticeable deviations in certain temporal segments.

\section{Limitations}
In this work, we introduce CARNet, an attention-free approach for jointly modeling recurrent cycles and cross-variate dependencies. Despite its effectiveness, CARNet has several limitations. First, similar to CycleNet~\cite{cyclenet} and TQNet~\cite{tqnet}, the cycle length $W$ is determined at the dataset level rather than adaptively for each sample, which may reduce effectiveness on datasets with weak, irregular, or non-stationary periodic patterns. Second, as CARNet is a channel-dependent model, its advantage is expected to be smaller on datasets with weak inter-variable correlations. Finally, although CARNet preserves linear-complexity interaction modeling, its performance still depends on appropriate architectural choices such as embedding size and network depth.

\section{Conclusion}
In this work, we proposed \textbf{CARNet}, an attention-free multivariate time series forecasting framework that extends core-based cross-variate modeling by incorporating explicit cycle conditioning. CARNet introduces a cycle-conditioned core aggregation and redistribution pipeline together with a \emph{Multihead Core Aggregation} (MHCA) mechanism, enabling structured and efficient interaction modeling across variates while preserving linear complexity. Unlike prior core-based methods that treat all variates uniformly, CARNet conditions both aggregation and redistribution on phase-aligned cycle representations, allowing the model to capture dataset-specific periodic structure in a principled manner. Empirical results on twelve real-world benchmarks demonstrate that CARNet consistently improves forecasting performance across multiple horizons while maintaining favorable computational efficiency. Ablation studies further confirm that each proposed component contributes meaningfully, with cycle-conditioned redistribution yielding the largest individual gain. These results highlight the benefit of integrating periodic structure into core-based architectures and suggest that explicit cycle conditioning is a promising direction for scalable multivariate time series forecasting.

%% Note that this preceding line implies that you store your BibTeX references in a file called 'mybibliography.bib'. If you instead store your references in a file with a different name, for instance 'references.bib', the preceding line should read '\bibliography{references}'. Whatever you do, DO NOT put the file name extension .bib inside the \bibliography command; this will trip up LaTeX compilers. 
%
% If you do not want to use BibTeX, you can also type up the bibliography exactly as you see fit, using the following structure:

\clearpage
% \bibliography{references}

\begin{thebibliography}{99}

\bibitem{patchtst}
Nie, Y., Ma, Q., Zhang, Q., Xu, Y., Wang, X.:
A Time Series is Worth 64 Words: Long-term Forecasting with Transformers.
In: Proc. Int. Conf. on Learning Representations (ICLR) (2023)

\bibitem{itransformer}
Liu, H., Lin, Z., Zhang, Y.:
iTransformer: Inverted Transformers Are Effective for Time Series Forecasting.
In: Proc. Int. Conf. on Learning Representations (ICLR) (2024)

\bibitem{instancenorm}
Ulyanov, D., Vedaldi, A., Lempitsky, V.:
Instance Normalization: The Missing Ingredient for Fast Stylization.
arXiv preprint arXiv:1607.08022 (2016)

\bibitem{timexer}
Woo, G., Kim, S., Song, H.O.:
TimeXer: Learning Multi-scale Temporal Representations for Long-term Forecasting.
In: Proc. Int. Conf. on Learning Representations (ICLR) (2024)

\bibitem{bigbird}
Zaheer, M. et al.:
BigBird: Transformers for Longer Sequences.
In: Proc. Advances in Neural Information Processing Systems (NeurIPS) (2020)

\bibitem{informer}
Zhou, H. et al.:
Informer: Beyond Efficient Transformer for Long Sequence Time-Series Forecasting.
In: Proc. AAAI Conf. on Artificial Intelligence (AAAI) (2021)

\bibitem{mamba}
Gu, A., Dao, T.:
Mamba: Selective State Space Models.
In: Proc. Int. Conf. on Learning Representations (ICLR) (2024)

\bibitem{smamba}
Wang, Z. et al.:
Is Mamba Effective for Time Series Forecasting?
Neurocomputing (2025)

\bibitem{timepro}
Ma, X. et al.:
TimePro: Efficient Multivariate Long-term Time Series Forecasting with Variable- and Time-Aware Hyper-state.
In: Proc. Int. Conf. on Machine Learning (ICML) (2025)

\bibitem{autoformer}
Wu, H. et al.:
Autoformer: Decomposition Transformers with Auto-Correlation for Long-Term Series Forecasting.
In: Proc. NeurIPS (2021)

\bibitem{fedformer}
Zhou, T. et al.:
FEDformer: Frequency Enhanced Decomposed Transformer for Long-term Series Forecasting.
In: Proc. ICML (2022)


\bibitem{tsmixer}
Liu, Y. et al.:
TSMixer: An All-MLP Architecture for Time Series Forecasting.
arXiv preprint arXiv:2306.09364 (2023)

\bibitem{softs}
Zhang, X. et al.:
SOFTS: Series-Core Fusion Transformer for Multivariate Time Series Forecasting.
In: Proc. NeurIPS (2024)

\bibitem{gwn}
Wu, Z. et al.:
Graph WaveNet for Deep Spatial-Temporal Graph Modeling.
In: Proc. IJCAI (2019)




\bibitem{sparsetsf}
Lin, S. et al.:
SparseTSF: Modeling Long-term Time Series Forecasting with 1k Parameters.
In: Proc. ICML (2024)

\bibitem{cyclenet}
Lin, S. et al.:
CycleNet: Enhancing Time Series Forecasting through Modeling Periodic Patterns.
In: Proc. NeurIPS (2024)

\bibitem{tqnet}
Lin, S. et al.:
Temporal Query Network for Efficient Multivariate Time Series Forecasting.
In: Proc. ICML (2025)

\bibitem{dlinear}
Zeng, A. et al.:
Are Transformers Effective for Time Series Forecasting?
In: Proc. AAAI (2023)

\bibitem{timemixer}
Liu, Y. et al.:
TimeMixer: Decomposable Multiscale Mixing for Time Series Forecasting.
In: Proc. ICLR (2024)

\bibitem{timesnet}
Wu, H. et al.:
TimesNet: Temporal 2D-Variation Modeling for General Time Series Analysis.
In: Proc. AAAI (2023)

\bibitem{pytorch}
Paszke, A. et al.:
PyTorch: An Imperative Style, High-Performance Deep Learning Library.
In: Proc. NeurIPS (2019)

\bibitem{adam}
Kingma, D.P., Ba, J.:
Adam: A Method for Stochastic Optimization.
In: Proc. ICLR (2015)

\bibitem{crossformer}
Zhang, Y. et al.:
Crossformer: Transformer Utilizing Cross-Dimension Dependency for Multivariate Time Series Forecasting.
In: Proc. NeurIPS (2023)

\bibitem{tide}
Das, A. et al.:
TiDE: Long-term Forecasting with Time-series Dense Encoder.
In: Proc. ICML (2023)

\bibitem{scinet}
Liu, Z. et al.:
SCINet: Time Series Modeling and Forecasting with Sample Convolution and Interaction.
In: Proc. NeurIPS (2021)



\bibitem{lstnet}
Lai, G. et al.:
Modeling Long- and Short-Term Temporal Patterns with Deep Neural Networks.
In: Proc. SIGIR (2018)

\bibitem{tfb}
Qiu, X. et al.:
TFB: Towards Comprehensive and Fair Benchmarking of Time Series Forecasting Methods.
PVLDB (2024)

\bibitem{deep_tss_survey}
Wang, Y. et al.:
Deep Time Series Models: A Comprehensive Survey and Benchmark.
arXiv:2407.13278 (2024)

\bibitem{transformers_ts_survey}
Li, C. et al.:
Transformers in Time Series: A Survey.
In: Proc. IJCAI (2023)

\bibitem{kedgn}
Zhang, J. et al.:
Knowledge-Empowered Dynamic Graph Network for Irregularly Sampled Medical Time Series.
In: Proc. NeurIPS (2024)

\bibitem{linearattention}
Katharopoulos, A. et al.:
Transformers are RNNs: Fast Autoregressive Transformers with Linear Attention.
In: Proc. NeurIPS (2020)

\bibitem{stochasticpooling}
Zeiler, M.~D. and Fergus, R.,
\newblock “Stochastic Pooling for Regularization of Deep Convolutional Neural Networks,”
\newblock in \emph{Proceedings of the International Conference on Learning Representations (ICLR)}, 2013.


\bibitem{madsen_ts}
Madsen, H.:
Time Series Analysis.
CRC Press (2007)

\bibitem{gelu}
Hendrycks, D., Gimpel, K.:
Gaussian Error Linear Units (GELUs).
arXiv:1606.08415 (2016)
\end{thebibliography}

\appendix
\section{More Details on CARNet}
\label{app:Algorithm}
The overall architecture of CARNet is described in Algorithm~\ref{alg:CARNet}, while the proposed Multihead core aggregation mechanism is detailed in Algorithm~\ref{alg:core-extract}. The pseudocode formalizes the forward computation of CARNet, highlighting how cycle information is integrated with variate embeddings through Cycle-Conditioned Core Aggregation and redistribution across multiple layers. This formulation clarifies the interaction between global periodic representations and cross-variate dependencies, and illustrates how the proposed modules are seamlessly embedded within the SOFTS-style star-shaped framework.

\begin{algorithm}[h]
\caption{Pseudocode of CARNet}
\label{alg:CARNet}
\begin{algorithmic}[1]
\REQUIRE Variate embedding $\mathbf{X}' \in \mathbb{R}^{C \times d}$, phase-aligned cycle $\mathbf{Z} \in \mathbb{R}^{C \times T}$

\FOR{$\ell = 1$ \TO $L$}

    \STATE $\mathbf{Z}' \leftarrow \mathcal{L}_Z(\mathbf{Z})$ \hfill // Cycle projection, $\mathbf{Z}' \in \mathbb{R}^{C \times d}$

    \STATE $\mathbf{V} \leftarrow \mathcal{F}(\mathbf{X}', \mathbf{Z}')$ \hfill // Feature fusion (concat), $\mathbf{V}\in\mathbb{R}^{C \times 2d}$

    \STATE $\mathbf{V}' \leftarrow \psi\!\Big(\mathcal{L}_V\big(\psi(\mathbf{V})\big)\Big)$ \hfill // Cycle-conditioned variate embedding, $\mathbf{V}'\in\mathbb{R}^{C \times d}$

    \STATE $\boldsymbol{\theta} \leftarrow \mathrm{MHCA}(\mathbf{V}')$ \hfill // Core aggregation, $\boldsymbol{\theta}\in\mathbb{R}^{C \times d'}$

    \STATE $\boldsymbol{\theta}' \leftarrow \beta\!\left(\mathcal{P}(\boldsymbol{\theta})\right)$ \hfill // Pool + broadcast, $\boldsymbol{\theta}'\in\mathbb{R}^{C \times d'}$

    \STATE $\tilde{\mathbf{V}} \leftarrow \mathcal{F}(\mathbf{X}', \mathbf{Z}', \boldsymbol{\theta}')$ \hfill // Redistribution fusion, $\tilde{\mathbf{V}}\in\mathbb{R}^{C \times (2d+d')}$

    \STATE $\mathbf{G} \leftarrow \mathcal{L}_F\!\Big(\psi\big(\mathcal{L}_{\tilde{V}}(\tilde{\mathbf{V}})\big)\Big)$ \hfill // Redistributed features, $\mathbf{G}\in\mathbb{R}^{C \times d}$

    \STATE $\mathbf{O} \leftarrow \mathrm{FFN}(\mathbf{X}' + \mathbf{G})$ \hfill // Output representation, $\mathbf{O}\in\mathbb{R}^{C \times d}$

    \STATE $\mathbf{X}' \leftarrow \mathbf{O}$ \hfill // Update for next layer

\ENDFOR

\RETURN $\mathbf{X}'$
\end{algorithmic}
\end{algorithm}

\begin{algorithm}[h]
\caption{Pseudocode of Multihead Core Aggregation (MHCA)}
\label{alg:core-extract}
\begin{algorithmic}[1]
\REQUIRE Input $\mathbf{V}' \in \mathbb{R}^{C \times d}$, number of heads $h$

\STATE $\mathbf{H} \leftarrow \alpha\!\left(\mathcal{L}_H(\mathbf{V}')\right)$ 
\hfill // Head-wise decomposition, $\mathbf{H}\in\mathbb{R}^{C \times h \times d_{\text{head}}}$

\FOR{$i = 1$ \TO $h$}
    \STATE $\boldsymbol{\theta}_i \leftarrow \psi\!\left(\omega(\mathbf{H}_i)\right)$
    \hfill // Head-wise core extraction, $\boldsymbol{\theta}_i\in\mathbb{R}^{C \times d'_{\text{head}}}$
\ENDFOR

\STATE $\tilde{\boldsymbol{\theta}} \leftarrow \alpha\!\left([\boldsymbol{\theta}_1;\dots;\boldsymbol{\theta}_h]\right)$
\hfill // Merge heads, $\tilde{\boldsymbol{\theta}}\in\mathbb{R}^{C \times d'}$

\STATE $\boldsymbol{\theta} \leftarrow \mathcal{L}_{\theta}(\tilde{\boldsymbol{\theta}})$
\hfill // Unified Multihead core, $\boldsymbol{\theta}\in\mathbb{R}^{C \times d'}$

\RETURN $\boldsymbol{\theta}$
\end{algorithmic}
\end{algorithm}

\section{Dataset Description}
\label{app:det_data}
We evaluate our method on twelve widely used real-world multivariate time-series datasets spanning multiple application domains, including electricity consumption, energy generation, traffic flow, and weather forecasting. The benchmarks include the ETT datasets, PEMS traffic datasets, and the Electricity, Solar, Traffic, and Weather datasets.

\begin{table}[h]
\caption{Detailed dataset description with standardized train/validation/test splits following prior benchmark settings \cite{tqnet,timexer,itransformer,softs}.}
\label{tab:dataset_info}
\centering
\footnotesize
\setlength{\tabcolsep}{3.2pt}
\renewcommand{\arraystretch}{0.7}

\begin{threeparttable}
\resizebox{\columnwidth}{!}{
\begin{tabular}{l| c |c |c |c |c}
\toprule
Dataset & Domain & Interval & Channels & Timesteps & Dataset Split \\
\midrule
ETTh1 & Electricity & 1h & 7 & 14400 & (8545, 2881, 2881) \\
\midrule
ETTh2 & Electricity & 1h & 7 & 14400 & (8545, 2881, 2881) \\
\midrule
ETTm1 & Electricity & 15m & 7 & 57600 & (34465, 11521, 11521) \\
\midrule
ETTm2 & Electricity & 15m & 7 & 57600 & (34465, 11521, 11521) \\
\midrule
ECL & Electricity & 1h & 321 & 26304 & (18317, 2633, 5261) \\
\midrule
Solar & Energy & 10m & 137 & 52560 & (36601, 5161, 10417) \\
\midrule
Traffic & Transportation & 1h & 862 & 17544 & (12185, 1757, 3509) \\
\midrule
Weather & Weather & 10m & 21 & 52696 & (36792, 5271, 10540) \\
\midrule
PEMS03 & Transportation & 5m & 358 & 26208 & (15617, 5135, 5135) \\
\midrule
PEMS04 & Transportation & 5m & 307 & 16992 & (10172, 3375, 3375) \\
\midrule
PEMS07 & Transportation & 5m & 883 & 28224 & (16911, 5622, 5622) \\
\midrule
PEMS08 & Transportation & 5m & 170 & 17856 & (10690, 3548, 3548) \\
\bottomrule
\end{tabular}
}

\begin{tablenotes}
\footnotesize
\item[1]\hypertarget{fn:ett}{} \url{https://github.com/zhouhaoyi/ETDataset}
\item[2]\hypertarget{fn:ecl}{} \url{https://archive.ics.uci.edu/ml/datasets/ElectricityLoadDiagrams20112014}
\item[3]\hypertarget{fn:pems}{} \url{https://pems.dot.ca.gov}
\end{tablenotes}
\end{threeparttable}
\end{table}

Following standard experimental settings, the forecasting horizons for the PEMS datasets are selected from \{12, 24, 48, 96\}, while horizons of \{96, 192, 336, 720\} are used for the remaining datasets. As summarized in Table~\ref{tab:dataset_info}, the datasets also differ substantially in temporal resolution, dimensionality, total timesteps, and standardized training/validation/test splits, providing a comprehensive benchmark for long-term forecasting.

The \textbf{ETT} dataset\hyperlink{fn:ett}{\textsuperscript{1}} consists of four electricity transformer temperature subsets, including two hourly datasets (ETTh1, ETTh2) and two 15-minute datasets (ETTm1, ETTm2), each with seven variables. \textbf{Traffic} records hourly road occupancy rates collected from sensors on San Francisco freeways. \textbf{Electricity (ECL)}\hyperlink{fn:ecl}{\textsuperscript{2}} provides hourly electricity consumption data from 321 clients. \textbf{Weather} contains 21 meteorological variables recorded every 10 minutes in Germany. \textbf{Solar-Energy} includes 10-minute solar power generation data from 137 photovoltaic plants. \textbf{PEMS}\hyperlink{fn:pems}{\textsuperscript{3}} comprises 5-minute traffic flow measurements collected from California traffic networks.

\section{Experiment Details}
\label{app:det_imp}
All experiments are conducted on a single NVIDIA GeForce RTX 5060 GPU with 8 GB VRAM. Performance comparisons among different methods are evaluated using two primary metrics: Mean Squared Error (MSE) and Mean Absolute Error (MAE). The training, validation, and test splits are kept consistent with TQNet \cite{tqnet}. Specifically, the data are divided using a 6:2:2 ratio for the ETT and PEMS series datasets and a 7:1:2 ratio for the remaining datasets. CARNet is trained for 30 epochs on all datasets except ETTm1 and ETTm2, where the number of epochs is limited to 7, with a patience counter of 3 applied across all datasets. A fixed random seed of 2024 is used to ensure reproducibility.

\section{Additional Prediction Visualizations}
\begin{figure*}[!h]
\centering
\includegraphics[width=0.8\columnwidth]{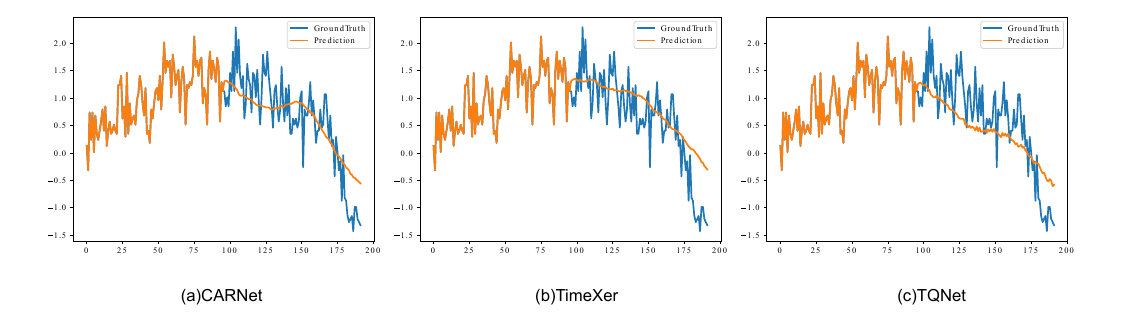}
\caption{Visualization of forecasting results on the PEMS08 dataset with look-back window 96 and prediction horizon 96.}
\label{PEMS08 prediction}
\end{figure*}

\begin{figure*}[!h]
\centering
\includegraphics[width=0.8\columnwidth]{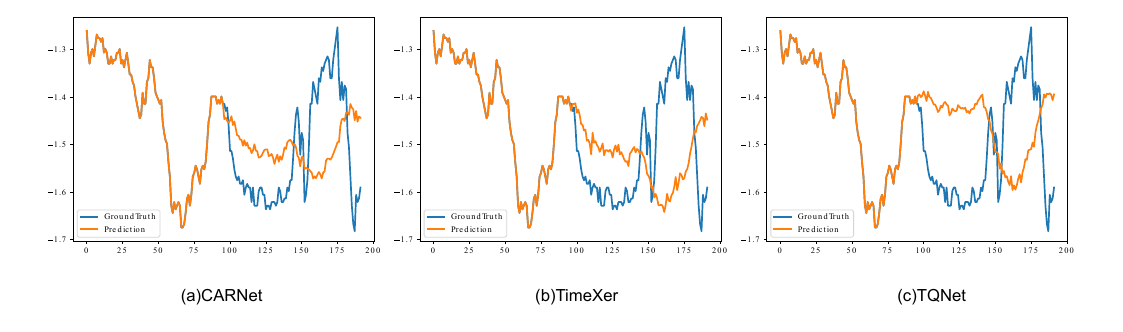}
\caption{Visualization of forecasting results on the ETTm1 dataset with look-back window 96 and prediction horizon 96.}
\label{ETTm1 prediction}
\end{figure*}

To further illustrate the forecasting behavior of CARNet, we provide additional qualitative prediction visualizations on two datasets, \textbf{PEMS08} and \textbf{ETTm1}. For both datasets, we consider an input sequence length of 96 and a prediction horizon of 96, and compare CARNet against representative strong baselines, including TimeXer and TQNet. Figures~\ref{PEMS08 prediction} and~\ref{ETTm1 prediction} present the corresponding prediction results on PEMS08 and ETTm1, respectively.

\section{Full Results of Component Ablation}
\label{app:full component}
To evaluate the contribution of individual components, we conduct ablation studies by removing each module from the full model and measuring the resulting performance degradation. Specifically, we remove (i) the Cycle-Conditioned Core Aggregation, which excludes the cycle projection and its fusion with the variate embeddings, (ii) the Multihead Core Aggregation module, and (iii) the Cycle-Conditioned Core Redistribution, which redistributes the cycle projection together with the cycle-conditioned core to the variate embeddings. Table~\ref{tab:component} reports the results across six datasets and four prediction horizons. The results indicate that each component contributes meaningfully to the overall forecasting performance.

\begin{table}[h]
\centering
\caption{Full results of the component ablation study.}
\label{tab:component}
\tiny
\setlength{\tabcolsep}{6pt}
\renewcommand{\arraystretch}{0.7}

\resizebox{1\columnwidth}{!}{
\begin{tabular}{c c | cc | cc | cc | cc}
\toprule
Dataset & Pred
& \multicolumn{2}{c}{cSOFTS}
& \multicolumn{2}{c}{w/o MHCA}
& \multicolumn{2}{c}{w/o CCCA}
& \multicolumn{2}{c}{w/o CCCR} \\
\cmidrule(lr){3-4}\cmidrule(lr){5-6}\cmidrule(lr){7-8}\cmidrule(lr){9-10}
& & MSE & MAE & MSE & MAE & MSE & MAE & MSE & MAE \\
\midrule

\multirow{5}{*}{ETTm1}
& 96  & \textbf{\textcolor{red}{0.304}} & \textbf{\textcolor{red}{0.348}} & {0.308} & {0.352} & 0.311 & 0.355 & 0.316 & 0.355 \\
& 192 & \textbf{\textcolor{red}{0.354}} & \textbf{\textcolor{red}{0.379}} & 0.357 & 0.381 & 0.359 & 0.382 & 0.361 & 0.380 \\
& 336 & \textbf{\textcolor{red}{0.388}} & \textbf{\textcolor{red}{0.401}} & 0.394 & 0.404 & 0.390 & 0.404 & 0.398 & 0.404 \\
& 720 & \textbf{\textcolor{red}{0.454}} & \textbf{\textcolor{red}{0.438}} & 0.460 & 0.443 & 0.457 & 0.441 & 0.459 & 0.439 \\
\midrule
& Avg
& \textbf{\textcolor{red}{0.375}} & \textbf{\textcolor{red}{0.392}}
& 0.380 & 0.395
& 0.379 & 0.396
& 0.384 & 0.395 \\

\midrule

\multirow{5}{*}{Solar}
& 96  & 0.177 & \textbf{\textcolor{red}{0.231}} & 0.178 & 0.245 & \textbf{\textcolor{red}{0.176}} & 0.237 & 0.189 & 0.238 \\
& 192 & \textbf{\textcolor{red}{0.198}} & \textbf{\textcolor{red}{0.250}} & 0.206 & 0.263 & 0.207 & 0.267 & 0.202 & 0.254 \\
& 336 & \textbf{\textcolor{red}{0.208}} & \textbf{\textcolor{red}{0.260}} & 0.213 & 0.268 & 0.222 & 0.274 & 0.236 & 0.282 \\
& 720 & \textbf{\textcolor{red}{0.209}} & \textbf{\textcolor{red}{0.256}} & 0.214 & 0.264 & 0.230 & 0.280 & 0.215 & 0.261 \\
\midrule
& Avg
& \textbf{\textcolor{red}{0.198}} & \textbf{\textcolor{red}{0.249}}
& 0.203 & 0.260
& 0.209 & 0.265
& 0.211 & 0.259 \\

\midrule

\multirow{5}{*}{Weather}
& 96  & \textbf{\textcolor{red}{0.155}} & \textbf{\textcolor{red}{0.200}} & 0.157 & 0.203 & 0.157 & 0.202 & 0.169 & 0.206 \\
& 192 & \textbf{\textcolor{red}{0.203}} & \textbf{\textcolor{red}{0.244}} & 0.207 & 0.247 & 0.205& 0.245 & 0.216 & 0.249 \\
& 336 & 0.265 & 0.289 & \textbf{\textcolor{red}{0.264}} & \textbf{\textcolor{red}{0.288}} & 0.263 & 0.290 & 0.276 & 0.294 \\
& 720 & \textbf{\textcolor{red}{0.344}} & \textbf{\textcolor{red}{0.342}} & \textbf{\textcolor{red}{0.344}} & \textbf{\textcolor{red}{0.342}} &{0.346} & 0.344 & 0.351 & 0.344 \\
\midrule
& Avg
& \textbf{\textcolor{red}{0.242}} & \textbf{\textcolor{red}{0.269}}
& 0.243 & 0.270
& 0.243 & 0.270
& 0.253 & 0.273 \\

\midrule

\multirow{5}{*}{PEMS04}
& 12  & \textbf{\textcolor{red}{0.064}} & \textbf{\textcolor{red}{0.161}} & 0.067 & 0.166 & 0.067 & 0.163 & 0.074 & 0.175 \\
& 24  & \textbf{\textcolor{red}{0.071}} & \textbf{\textcolor{red}{0.170}} & 0.077 & 0.167 & 0.075 & 0.172 & 0.089 & 0.194 \\
& 48  & \textbf{\textcolor{red}{0.083}} & \textbf{\textcolor{red}{0.185}} & 0.086 & 0.187 & 0.085 & 0.186 & 0.112 & 0.220 \\
& 96  & \textbf{\textcolor{red}{0.099}} & \textbf{\textcolor{red}{0.203}} & 0.102 & 0.206 & 0.100 & 0.205 & 0.138 & 0.247 \\
\midrule
& Avg
& \textbf{\textcolor{red}{0.079}} & \textbf{\textcolor{red}{0.180}}
& 0.083 & 0.182
& 0.082 & 0.182
& 0.103 & 0.209 \\

\midrule

\multirow{5}{*}{PEMS08}
& 12  & \textbf{\textcolor{red}{0.069}} & \textbf{\textcolor{red}{0.164}} & \textbf{\textcolor{red}{0.069}} & 0.165 & \textbf{\textcolor{red}{0.069}} & 0.165 & 0.076 & 0.173 \\
& 24  & \textbf{\textcolor{red}{0.087}} & \textbf{\textcolor{red}{0.185}} & 0.090 & 0.187 & 0.092 & 0.188 & 0.105 & 0.202 \\
& 48  & \textbf{\textcolor{red}{0.125}} & \textbf{\textcolor{red}{0.212}} & 0.128 & 0.213 & 0.132 & 0.222 & 0.167 & 0.253 \\
& 96  & \textbf{\textcolor{red}{0.204}} & \textbf{\textcolor{red}{0.250}} & 0.234 & 0.262 & 0.206 & 0.248 & 0.265 & 0.320 \\
\midrule
& Avg
& \textbf{\textcolor{red}{0.121}} & \textbf{\textcolor{red}{0.203}}
& 0.130 & 0.207
& 0.125 & 0.206
& 0.153 & 0.237 \\

\midrule

\multirow{5}{*}{ECL}
& 96  & \textbf{\textcolor{red}{0.131}} & \textbf{\textcolor{red}{0.226}} & 0.132 & \textbf{\textcolor{red}{0.226}} & 0.131 & \textbf{\textcolor{red}{0.226}} & 0.147 & 0.237 \\
& 192 & \textbf{\textcolor{red}{0.149}} & \textbf{\textcolor{red}{0.244}} & 0.152 & 0.247 & 0.151 & \textbf{\textcolor{red}{0.244}} & 0.160 & 0.251 \\
& 336 & \textbf{\textcolor{red}{0.164}} & \textbf{\textcolor{red}{0.260}} & 0.168 & 0.264 & 0.165 & 0.262 & 0.178 & 0.270 \\
& 720 & \textbf{\textcolor{red}{0.188}} & \textbf{\textcolor{red}{0.285}} & 0.194 & 0.288 & 0.192 & 0.286 & 0.208 & 0.297 \\
\midrule
& Avg
& \textbf{\textcolor{red}{0.158}} & \textbf{\textcolor{red}{0.254}}
& 0.162 & 0.256
& 0.160 & 0.255
& 0.173 & 0.264 \\

\bottomrule
\end{tabular}
}
\end{table}

\section{Results Under Different Random Seeds}
To further examine the robustness of CARNet, we conduct experiments on five representative datasets using multiple random initializations. For each dataset, results are reported across three random seeds for four forecasting horizons, and we summarize performance using the mean and standard deviation. As presented in Table~\ref{tab:seed_results}, CARNet consistently achieves low standard deviation across a wide range of settings, indicating stable training dynamics and reliable forecasting performance. Overall, these results demonstrate that the effectiveness of CARNet is not dependent on a particular random seed and generalizes consistently across datasets with diverse temporal characteristics.

\begin{table}[!h]
\caption{Performance of CARNet under different random seeds. Mean represents the average value, and Std denotes the standard deviation.}
\label{tab:seed_results}
\scriptsize
\setlength{\tabcolsep}{5pt}
\renewcommand{\arraystretch}{0.7}
`
\resizebox{1\columnwidth}{!}{
\begin{tabular}{c c | cc | cc | cc | cc | cc}
\toprule
\multirow{2}{*}{Dataset} & \multirow{2}{*}{Pred} &
\multicolumn{10}{c}{Random Seed} \\
\cmidrule(lr){3-12}
& &
\multicolumn{2}{c}{2024} &
\multicolumn{2}{c}{2025} &
\multicolumn{2}{c}{2026} &
\multicolumn{2}{c}{Mean} &
\multicolumn{2}{c}{Std} \\
\cmidrule(lr){3-4} \cmidrule(lr){5-6} \cmidrule(lr){7-8}
\cmidrule(lr){9-10} \cmidrule(lr){11-12}
& & MSE & MAE & MSE & MAE & MSE & MAE & MSE & MAE & MSE & MAE \\
\midrule

% ================= Electricity =================
\multirow{4}{*}{Electricity}
& 96  & 0.131 & 0.226 & 0.132 & 0.228 & 0.131 & 0.227 & 0.1313 & 0.2270 & 0.0005 & 0.0008 \\
& 192 & 0.151 & 0.244 & 0.150 & 0.245 & 0.150 & 0.245 & 0.1503 & 0.2447 & 0.0005 & 0.0005 \\
& 336 & 0.166 & 0.260 & 0.166 & 0.265 & 0.165 & 0.264 & 0.1657 & 0.2630 & 0.0005 & 0.0022 \\
& 720 & 0.190 & 0.285 & 0.192 & 0.289 & 0.198 & 0.293 & 0.1933 & 0.2890 & 0.0034 & 0.0033 \\
\midrule

% ================= Weather =================
\multirow{4}{*}{Weather}
& 96  & 0.155 & 0.200 & 0.155 & 0.200 & 0.155 & 0.200 & 0.1550 & 0.2000 & 0.0000 & 0.0000 \\
& 192 & 0.203 & 0.244 & 0.203 & 0.245 & 0.205 & 0.245 & 0.2037 & 0.2447 & 0.0009 & 0.0005 \\
& 336 & 0.265 & 0.289 & 0.267 & 0.292 & 0.265 & 0.289 & 0.2657 & 0.2900 & 0.0009 & 0.0014 \\
& 720 & 0.344 & 0.342 & 0.350 & 0.344 & 0.345 & 0.343 & 0.3463 & 0.3430 & 0.0026 & 0.0008 \\
\midrule

% ================= Solar =================
\multirow{4}{*}{Solar}
& 96  & 0.177 & 0.231 & 0.185 & 0.243 & 0.190 & 0.241 & 0.1840 & 0.2383 & 0.0053 & 0.0053 \\
& 192 & 0.198 & 0.250 & 0.200 & 0.251 & 0.210 & 0.265 & 0.2027 & 0.2553 & 0.0052 & 0.0071 \\
& 336 & 0.208 & 0.260 & 0.213 & 0.261 & 0.204 & 0.257 & 0.2083 & 0.2593 & 0.0037 & 0.0017 \\
& 720 & 0.209 & 0.256 & 0.212 & 0.264 & 0.236 & 0.282 & 0.2190 & 0.2673 & 0.0121 & 0.0109 \\
\midrule

% ================= PEMS04 =================
\multirow{4}{*}{PEMS04}
& 12 & 0.064 & 0.161 & 0.064 & 0.161 & 0.063 & 0.161 & 0.0637 & 0.1610 & 0.0005 & 0.0000 \\
& 24 & 0.071 & 0.170 & 0.070 & 0.169 & 0.072 & 0.170 & 0.0710 & 0.1697 & 0.0008 & 0.0005 \\
& 48 & 0.083 & 0.185 & 0.084 & 0.187 & 0.085 & 0.188 & 0.0840 & 0.1867 & 0.0008 & 0.0012 \\
& 96 & 0.099 & 0.203 & 0.097 & 0.202 & 0.097 & 0.202 & 0.0977 & 0.2023 & 0.0009 & 0.0005 \\
\midrule

% ================= PEMS08 =================
\multirow{4}{*}{PEMS08}
& 12 & 0.069 & 0.164 & 0.069 & 0.166 & 0.069 & 0.164 & 0.0690 & 0.1647 & 0.0000 & 0.0009 \\
& 24 & 0.087 & 0.185 & 0.092 & 0.188 & 0.092 & 0.191 & 0.0903 & 0.1880 & 0.0024 & 0.0024 \\
& 48 & 0.125 & 0.212 & 0.129 & 0.214 & 0.129 & 0.214 & 0.1277 & 0.2133 & 0.0019 & 0.0009 \\
& 96 & 0.204 & 0.250 & 0.225 & 0.273 & 0.208 & 0.252 & 0.2123 & 0.2583 & 0.0092 & 0.0104 \\
\bottomrule
\end{tabular}
}
\end{table}

\end{document}